\UseRawInputEncoding
\documentclass[final,5p,times,twocolumn]{elsarticle}
\usepackage[utf8]{inputenc}

\usepackage{graphicx}
\usepackage{amssymb}
\usepackage{lineno}
\usepackage{natbib}
\usepackage{graphicx, changepage}
\usepackage{amsfonts}
\usepackage{amsmath, amssymb}
\usepackage{bm}
\usepackage{ctable}
\usepackage{booktabs}
\usepackage{multirow}
\usepackage{lscape}
\usepackage{float}
\usepackage{tabularx}
\usepackage{caption}
\usepackage[utf8]{inputenc}
\usepackage{listings}
\usepackage{chngcntr}
\usepackage{xcolor}
\usepackage{epstopdf}
\AppendGraphicsExtensions{.tif}
\usepackage[ruled,vlined]{algorithm2e}

\definecolor{codegreen}{rgb}{0,0.6,0}
\definecolor{codegray}{rgb}{0.5,0.5,0.5}
\definecolor{codepurple}{rgb}{0.58,0,0.82}
\definecolor{backcolour}{rgb}{0.95,0.95,0.92}

\definecolor{codegreen}{rgb}{0,0.6,0}
\definecolor{codegray}{rgb}{0.5,0.5,0.5}
\definecolor{codepurple}{rgb}{0.58,0,0.82}
\definecolor{backcolour}{rgb}{0.95,0.95,0.92}

\lstdefinestyle{mystyle}{
    backgroundcolor=\color{backcolour},   
    commentstyle=\color{codegreen},
    keywordstyle=\color{magenta},
    numberstyle=\tiny\color{codegray},
    stringstyle=\color{codepurple},
    basicstyle=\ttfamily\footnotesize,
    breakatwhitespace=false,         
    breaklines=true,                 
    captionpos=b,                    
    keepspaces=true,                 
    numbers=left,                    
    numbersep=5pt,                  
    showspaces=false,                
    showstringspaces=false,
    showtabs=false,                  
    tabsize=2
}

\newcommand{\abbreviations}[1]{%
  \nonumnote{\textit{Abbreviations:\enspace}#1}}

\biboptions{semicolon,round,sort,authoryear}
\renewcommand{\lstlistingname}{\bfseries Listing}
\makeatletter
\def\fnum@lstlisting{%
  \lstlistingname
  \ifx\lst@@caption\@empty\else~\thelstlisting\normalfont\fi}%
\makeatother

\makeatletter
    \AtBeginDocument{
    \renewcommand*{\thelstlisting}{F.\arabic{lstlisting}}
    \@addtoreset{lstlisting}{section}}
\makeatother

\DeclareMathOperator*{\argmin}{arg\,min}

\begin{document}

\begin{frontmatter}

\title{Decomposing Normal and Abnormal Features of Medical Images into \\ Discrete Latent Codes for Content-Based Image Retrieval}

\author[a,b]{Kazuma Kobayashi\corref{correspondingauthor}}
\cortext[correspondingauthor]{Corresponding to: Kazuma Kobayashi, M.D., D.Sc. {\it Postal address:} Division of Medical AI Research and Development, National Cancer Center Research Institute, 5-1-1 Tsukiji Chuo-ku, Tokyo, 104-0045, Japan; {\it Email address:} {\texttt kazumkob@ncc.go.jp}}
\ead{kazumkob@ncc.go.jp}

\author[a,c]{Ryuichiro Hataya}
\ead{hataya@nlab.ci.i.u-tokyo.ac.jp}
\author[d,e]{Yusuke Kurose}
\ead{kurose@mi.t.u-tokyo.ac.jp}
\author[f]{Mototaka Miyake}
\ead{mmiyake@ncc.go.jp}
\author[g]{\\Masamichi Takahashi}
\ead{masataka@ncc.go.jp}
\author[h]{Akiko Nakagawa}
\ead{7bmm1342@cc.u-tokai.ac.jp}
\author[d,e]{Tatsuya Harada}
\ead{harada@mi.t.u-tokyo.ac.jp}
\author[a,b]{Ryuji Hamamoto}
\ead{rhamamot@ncc.go.jp}

\address[a]{Division of Medical AI Research and Development, National Cancer Center Research Institute, \\5-1-1 Tsukiji, Chuo-ku, Tokyo 104-0045, Japan}
\address[b]{Cancer Translational Research Team, RIKEN Center for Advanced Intelligent Project, \\1-4-1 Nihonbashi, Chuo-ku, Tokyo 103-0027, Japan}
\address[c]{Graduate School of Information Science and Technology, The University of Tokyo, \\7-3-1 Hongo, Bunkyo-ku, Tokyo 113-8656, Japan}
\address[d]{Research Center for Advanced Science and Technology, The University of Tokyo, \\4-6-1 Komaba, Meguro-ku, Tokyo 153-8904, Japan}
\address[e]{Machine Intelligence for Medical Engineering Team, RIKEN Center for Advanced Intelligent Project, \\1-4-1 Nihonbashi, Chuo-ku, Tokyo 103-0027, Japan}
\address[f]{Department of Diagnostic Radiology, National Cancer Center Hospital, \\5-1-1 Tsukiji, Chuo-ku, Tokyo 104-0045, Japan}
\address[g]{Department of Neurosurgery and Neuro-Oncology, National Cancer Center Hospital, \\5-1-1 Tsukiji, Chuo-ku, Tokyo 104-0045, Japan}
\address[h]{Tokai University School of Medicine, \\143 Shimokasuya, Isehara-shi, Kanagawa 259-1193, Japan}

\begin{abstract}
In medical imaging, the characteristics purely derived from a disease should reflect the extent to which abnormal findings deviate from the normal features. Indeed, physicians often need corresponding images without abnormal findings of interest or, conversely, images that contain similar abnormal findings regardless of normal anatomical context. This is called comparative diagnostic reading of medical images, which is essential for a correct diagnosis. To support comparative diagnostic reading, content-based image retrieval (CBIR), which can selectively utilize normal and abnormal features in medical images as two separable semantic components, will be useful. Therefore, we propose a neural network architecture to decompose the semantic components of medical images into two latent codes: \emph{normal anatomy code} and \emph{abnormal anatomy code}. The normal anatomy code represents normal anatomies that should have existed if the sample is healthy, whereas the abnormal anatomy code attributes to abnormal changes that reflect deviation from the normal baseline. These latent codes are discretized through vector quantization to enable binary hashing, which can reduce the computational burden at the time of similarity search. By calculating the similarity based on either normal or abnormal anatomy codes or the combination of the two codes, our algorithm can retrieve images according to the selected semantic component from a dataset consisting of brain magnetic resonance images of gliomas. Our CBIR system qualitatively and quantitatively achieves remarkable results. 
\end{abstract}


\abbreviations{2D, two-dimensional; 3D, three-dimensional; CBIR, content-based image retrieval; DM, distribution matching; ED, peritumoral edema; ET, Gd-enhancing tumor; FDN, feature decomposing network; FLAIR, fluid-attenuated inversion recovery; GAN, generative adversarial network; MR, magnetic resonance; NET, necrotic and non-enhancing tumor core; PPV, positive predictive value; SPADE, spatially adaptive normalization; UNIT, unsupervised image-to-image translation; VAE, variational autoencoder; VQ-VAE, vector-quantized variational autoencoder}

\end{frontmatter}

\section{Introduction}
\label{sec:introduction}

Abnormalities in medicine can be viewed as deviations from a normal feature in a healthy population. In medical images, such abnormalities appear as local or global deviations from an original normal anatomy. These changes should reflect a process where abnormalities occur under some initiation mechanisms from the normal anatomy and, with disease progression, existing normal tissues are eventually replaced by pathological ones. Indeed, when evaluating images of conditions to be diagnosed, physicians often need corresponding images without abnormal findings of interest or, conversely, images that contain similar abnormal findings regardless of normal anatomical context. This is called comparative diagnostic reading of medical images, which is essential for a correct diagnosis. For example, by comparing abnormal images of interest with normal images of the same anatomical site, physicians can diagnose the presence and extent of the disease. Meanwhile, images with similar abnormal features can also be useful as a reference because diseases with the same imaging phenotypes often have similar prognoses and treatment responses, somewhat regardless of the origin of the diseases. However, it is quite laborious to find similar images from a large database by focusing only on either normal or abnormal features of a query image by hands. Hence, we aimed to establish a content-based image retrieval (CBIR) system to support comparative diagnostic reading. 

CBIR is an important application to retrieve a set of the most similar images to a submitted patient's image from large databases to support clinical decision making. A CBIR system basically comprises two subsystems: At the first feature extraction stage, it converts images in a database to a set of features associated with image content, and at the next similarity search stage, similar images are retrieved from the database based on the feature similarity with respect to a query image. Traditionally, the feature extraction employed various types of handcrafted descriptors, such as shape, text, color, and texture; however, there was difficulty in reducing the ``semantic gap'' between low-level image features captured by the handcrafted descriptors and high-level visual concepts \citep{Kumar2013, sift7935507}. Recently, with the success in many image processing tasks, the deep-learning-based approach has gained increasing attention in the field of CBIR. Since the deep neural network can automatically learn complex features in hierarchical levels of abstraction, it can simplify the feature extraction process. A common neural network architecture used in CBIR systems is the autoencoder, which enables to map each image into a latent representation that can store compressed image information \citep{7727562, 7172448, 8310624, 8079843, 7351389}. Even though deep-learning-based approach often outperformed classical algorithm of CBIR systems \citep{10.1117/12.2251115, 10.1117/12.2217587}, there is no specific method that can measure the distance between either normal or abnormal features of medical images as separated semantic components. To achieve our goals, we need to devise a novel feature extraction method that can decompose normal and abnormal features of medical images. We also aimed to establish a computationally effective method to retrieve similar images based on decomposed latent representations.

\begin{figure}
  \centering
  \includegraphics[]{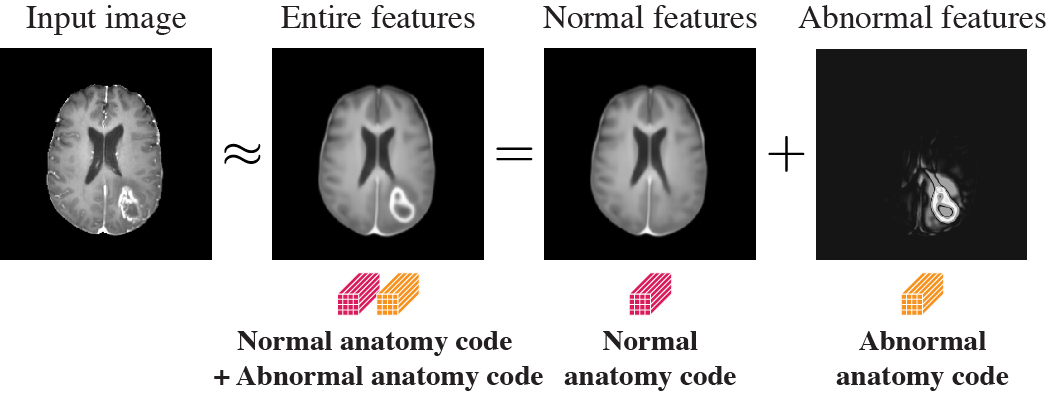}
  \caption{\textbf{Concept of the compositionality of medical images.} It indicates that the entire medical image can be decomposed into normal and abnormal anatomy codes relevant to normal and abnormal semantic components in the image, respectively.}
  \label{fig:concept_of_decomp}
\end{figure}

In this study, we define the two-tiered semantic nature of normal and abnormal anatomy as \emph{compositionality} of medical images, as presented in {\bf Fig. \ref{fig:concept_of_decomp}}. Hereinafter, ``normal'' anatomy means counterfactual structures that should have existed if the sample is healthy, whereas ``abnormal'' anatomy attributes to disease changes that reflect the deviation from the normal baseline. When the sample is free from abnormality, ``normal'' anatomy corresponds to the entire sample image, and ``abnormal'' anatomy should not indicate any condition. Then, we consider how to decompose a given image into two low-dimensional representations in a manipulable manner, where the latent codes representing normal and abnormal anatomy should be mutually exclusive and collective in terms of the reconstruction of the original image. By measuring similarities based on decomposed normal or abnormal semantic components of medical images, it can be expected to retrieve images that have the same normal anatomical context or similar abnormal findings, respectively, while ignoring the other component.

Factorizing compositionality into distinct and informative factors in data is a fundamental requirement of representation learning \citep{Bengio2013}. The majority of studies has sought to capture purely independent factors of variation that contributes to generation of data, which are called \emph{disentangled representations} \citep{higgins2018definition}. To obtain disentanglement of features, two approaches have exploited implicit or explicit supervision to impose strong inductive biases on autoencoders \citep{tschannen2018recent}. One approach is to collect a large amount of data to encompass sufficient variation for each factor and then apply appropriate regularization such that disentanglement can be implicitly performed \citep{Higgins2017betaVAELB, burgess2018understanding, pmlr-v80-kim18b, Chen2018tcvae, zhao2017infovae, kumar2017variational, lopez2018information, esmaeili2018structured, Alex2017, Achille2018, Lample2017, louizos2015variational}. Another approach is to force a model to acquire separate representations by explicitly imposing modeling assumptions \citep{cheung2014discovering, Kulkarni2015, Eslami2016attend, shanahan2019explicitly, charakorn2020explicit}. If distinct factors of variations explaining the characteristics of data can be separately represented using independent latent units, it can be useful in downstream tasks by providing interpretability and manipulability to the data. 

Therefore, disentangled representations can decompose normal and abnormal semantic components of medical images. Indeed, there has been an increasing number of studies focusing on feature disentanglement in medical imaging, including performing pseudo-healthy synthesis \citep{XIA2020101719, ADN8788607, vorontsov2019semisupervised, Tang2021}, learning generalizable features across domains \citep{meng2020learning, 9247170} or imaging modalities \citep{Chen2019, 9250615, Chartsias2019}, and evaluating the reliability of individual annotators from true segmentation label distributions \citep{HumanError2020}. However, with respect to the etiology of diseases in medical images, it is noteworthy that the assumption of disentanglement, where purely independent low-dimensional latent features can mimic the generation of high-dimensional data, can be too simple to be generalizable. For example, brain tumors originate as focal changes, and with their progression, cause compressional deformations and infiltration in adjacent structures, making it difficult to clearly define the boundary between normal and abnormal tissues. Hence, we consider \emph{decomposition} of latent representation, which encompasses the notion of disentangled representation as its special case and allows much richer properties of latent spaces such as intricate structured relationships \citep{pmlr-v97-mathieu19a}. 

In this study, we first devised a neural network architecture, called \emph{feature decomposing network}, to decompose normal and abnormal semantic components of medical images in a manipulable manner. Then, we demonstrated a novel CBIR framework by utilizing the decomposed latent codes to support comparative diagnostic reading. Given an input image, the feature decomposing network is trained to map it into two latent codes, {\it normal anatomy code} and {\it abnormal anatomy code}. It indicates that the original latent space for representing the whole image is divided into one portion as a normal anatomy code corresponding to normal anatomy and the remaining portion as an abnormal anatomy code corresponding to abnormal anatomy. After training the feature decomposing network, latent codes become representative of targeted semantic features in medical images. We also investigate the effectiveness of a method called \emph{distribution matching} by utilizing Wasserstein generative adversarial networks (GANs) \citep{pmlr-v70-arjovsky17a} with gradient penalty \citep{NIPS2017_892c3b1c}. Distribution matching is imposed on the latent distribution to minimize the overlap in the semantic contents between normal and abnormal anatomy codes. Furthermore, by constructing the two latent codes to be discrete through vector quantization \citep{oord2017neural}, we can reduce computational burden at the time of similarity search by binary hashing. The utility of these decomposed latent codes for CBIR applications is shown based on a large dataset containing brain magnetic resonance (MR) images of gliomas. By performing nearest neighbor search utilizing either normal or abnormal anatomy codes or the combination of the two codes, our CBIR system can retrieve images according to selected semantic components while ignoring the other, if necessary.

The main contributions of this study can be summarized into the following:
\begin{itemize}
  \item We propose a feature decomposing network that can explicitly decompose semantic features of medical images into normal and abnormal anatomy codes in a manipulable manner for downstream tasks. 
  \item To enhance computational efficiency at the time of similarity search, latent spaces are configured using vector quantization to be discrete, rather than continuous.
  \item By employing the decomposed latent codes, we present a novel CBIR application that can search for similarities in images based on a selected latent code or the combination of the two latent codes, enabling retrieval of images viewed as any semantic components while ignoring the other, if necessary. 
\end{itemize}

The proposed method is most closely related to our conference extended abstract \citep{kobayashi2020decomposing}, a neural network architecture to decompose normal and abnormal features of medical images with its application to CBIR. However, the presented work has significantly improved the learning method with extensive experimental settings. We further validated the performance of the CBIR application with more appropriate metrics. 

The rest of the manuscript is organized as follows: {\bf Section \ref{sec:related_work}} reviews the literature on disentangled representation, image-to-image translation in medical imaging, and CBIR. {\bf Section \ref{sec:methodology}} describes our proposed method with technical backgrounds. {\bf Section \ref{sec:experiments}} presents experimental settings and evaluation methods. {\bf Section \ref{sec:results}} provides the results. {\bf Section \ref{sec:discussion}} presents the discussion and conclusions. 

\section{Related work}
\label{sec:related_work}

Here, we briefly review literature related to disentangled representation learning, especially in the field of computer vision. Thereafter, as a research interest more related to our purpose, we review recent progress that mainly apply the image-to-image translation technique based on GANs for pseudo-healthy synthesis of medical images. We also introduce current progress in the CBIR system in medical imaging. 

\subsection{Disentangled representation learning}
\label{sec:disentangleed_representation_learning}
Learning disentangled representation attempts to separate the underlying factors of sample variations in a way that each factor exclusively represents one type of sample attributes. There are several benefits in learning disentangled representation from data because models that produce feature disentanglement can provide explainability of the model function and manipulability in the data generation process. One approach is to combine deep generative models, such as GANs \citep{Goodfellow2014} and variational autoencoders (VAEs) \citep{Kingma2013, Rezende2014} with regularization techniques to acquire disentanglement in an implicit manner \citep{Higgins2017betaVAELB, pmlr-v80-kim18b, Chen2018tcvae, zhao2017infovae, kumar2017variational, lopez2018information, esmaeili2018structured, Alex2017, Achille2018, Lample2017, louizos2015variational}. For example, $\beta$-VAE can automatically discover the independent latent factors of variation by imposing a limit on the capacity of latent information, which facilitates factorization of representations \citep{Higgins2017betaVAELB}. However, it is still fundamentally difficult to learn disentangled features without any supervision. It is also argued that acquired disentangled representations sometimes mismatch human's predefined concepts \citep{Locatello}. Another approach is to explicitly factorize representations into a component that is related to or independent of classes based on labeled data \citep{cheung2014discovering, Kulkarni2015, Eslami2016attend, shanahan2019explicitly, charakorn2020explicit}. Label information or attribute annotation can serve as strong supervision for feature disentanglement; hence, the performance of disentanglement can be optimized and guaranteed. Therefore, by designing a network with an appropriate structure and exploiting segmentation labels indicating abnormal regions as supervision, we aim to obtain the desired decomposition of latent representations. 

\subsection{Image-to-image translation}
\label{sec:image_to_image_translation}
Image-to-image translation has been exploited in medical imaging to obtain disentangled representation. CycleGAN, which performs bidirectional translation between image domains, has been widely used in image-to-image translation \citep{cyclegan8237506}. As an extended architecture of CycleGAN, the unsupervised image-to-image translation (UNIT) framework proposes a shared latent space assumption, where a pair of corresponding images in two different domains can be mapped \citep{liu2017unsupervised}. More recently, multimodal UNIT decomposes an image into a content code that is domain invariant and a style code that represents domain-specific features \citep{huang2018multimodal}. In line with studies focusing on medical imaging, Xia et al. demonstrated pseudo-healthy synthesis by creating a subject-specific healthy image from a pathological one by extending the learning framework of CycleGAN \citep{XIA2020101719}. Similarly, Liao et al. proposed an artifact disentanglement network using the image-to-image translation architecture, achieving comparable performance in image restoration to existing supervised models \citep{ADN8788607}. Vorontsov et al. also applied the same concept to improve semi-supervised training for semantic segmentation with autoencoding \citep{vorontsov2019semisupervised}. To enhance realistic synthesis of chest radiographic images, Tang et al. proposed a disentangled generative model for disease decomposition and demonstrated that disease residual maps can indicate underlying abnormal regions \citep{Tang2021}. Since one of our goals is to decompose medical images into normal and abnormal semantic components, the basic idea is somewhat similar to pseudo-healthy synthesis exploiting image-to-image translation techniques. However, previous approaches focused on transforming rather superficial appearance of images and did not evaluate the accessibility and validity of latent representations acquired inside the models. Thus, our feature decomposing network has a bottleneck where imaging features are compressed, enabling to handle latent representations of targeted semantic component for the downstream task. 

\subsection{Content-based image retrieval}
\label{sec:content_based_image_retrieval}
CBIR is an important application in retrieving a set of the most similar images from large databases, given a query image. Similarity measurements based on various information, such as shape, text, color and texture, and features acquired inside convolutional neural networks are used to resolve the semantic gap between imaging features and high-level visual concepts \citep{Kumar2013, sift7935507}. Even though several studies utilizing state-of-the-art techniques of deep neural networks have been introduced to CBIR \citep{Haq2021, MohdZin2018}, to the best of our knowledge, there are few studies that exploit disentangled representation from the viewpoint of image retrieval. Recently, Havaei et al. proposed a neural network architecture to ensure content-style disentanglement of medical images and demonstrated how the inferred style and content features are disentangled from each other by utilizing CBIR as an evaluation method \citep{havaei2020conditional}. Since CBIR is one of essential technologies that assist physicians in diagnosis, a methodology that directly seeks to develop CBIR based on disentangled representation is worth considering. Therefore, in addition to the concept of decomposing normal and abnormal features of medical images, we particularly employ latent codes as to be discrete through vector quantization. By applying vector quantization for the latent space, subspaces can be fixed and transversed by the Hamming distance rearrangement, which is favorable in a large-scale CBIR by reducing the computational cost at the time of similarity search \citep{9050998}. 

\section{Material and methods}
\label{sec:methodology}

In this study, we propose a method to decompose two-tiered semantic components of medical images into normal and abnormal anatomical codes for the application of CBIR that can selectively focus on semantic components. The proposed method is presented in two stages: First, we describe a network architecture, which we call \emph{feature decomposing network}, to decompose normal and abnormal features in medical images and its learning strategy. Then, we present how to establish a CBIR system that can support comparative diagnostic reading by utilizing these decomposed features. 

\subsection{Feature decomposing network}
\label{sec:training_of_feature_decomposing_network}

\begin{figure*}
  \centering
  \includegraphics[]{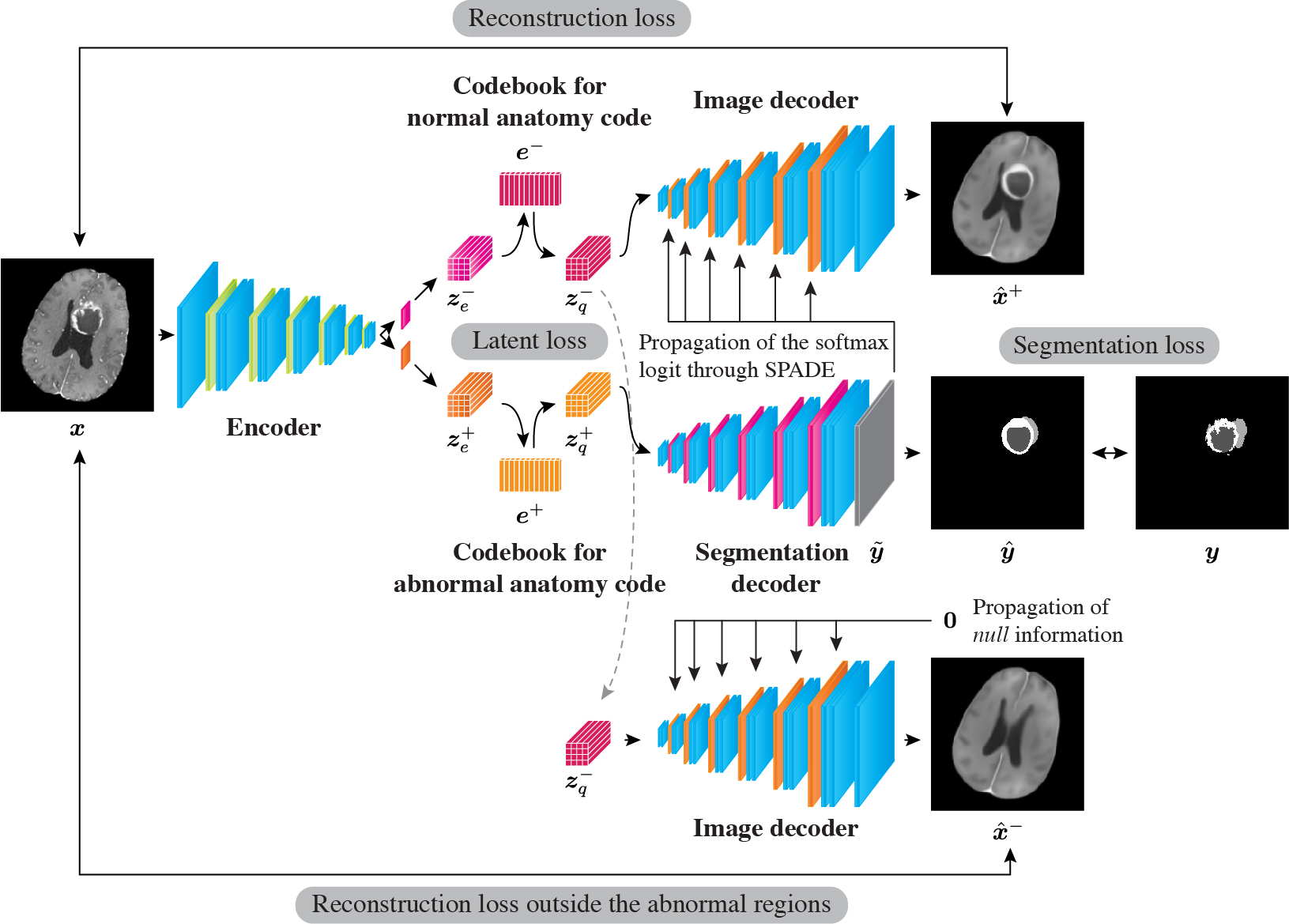}
  \caption{\textbf{Shared architecture of the feature decomposing networks.} Input image $\bm{x}$ is mapped to a pair of latent representations, $\bm{z}_e^-$ and $\bm{z}_e^+$. Vector quantization is performed based on two codebooks, $\bm{e}^-$ and $\bm{e}^+$, to produce normal anatomy code $\bm{z}_q^-$ and abnormal anatomy code $\bm{z}_q^+$, respectively. The segmentation decoder predicts the semantic segmentation of abnormality $\hat{\bm{y}}$ from $\bm{z}_q^+$. Depending on the collateral input of the softmax logit of abnormality $\tilde{\bm{y}}$, the image decoder reconstructs either the entire input image $\hat{\bm{x}}^+$ or normal-appearing image ${\hat{\bm{x}}^-}$. Several loss functions in training the network, including latent, reconstruction and segmentation losses, are shown, except for those related to distribution matching.}
  \label{fig:fdn_model_architecture}
\end{figure*}

The feature decomposing network is used to decompose semantic components of medical images into normal and abnormal anatomy codes. It is trained based on a dataset containing input images $\bm{x}$ and ground-truth segmentation labels $\bm{y}$. The feature decomposing network consists of an encoder and two decoders, segmentation decoder, and image decoder. A pair of discrete latent spaces with latent codebooks exists at the bottom of the network, each of which produces normal and abnormal anatomy codes. The overview of the network architecture is shown in {\bf Fig. \ref{fig:fdn_model_architecture}}. It is noteworthy that the feature decomposing network has the latent space as its bottleneck, where no bypass connection between the encoder and decoders, such as skip connections \citep{drozdzal2016importance}, is implemented. Therefore, we can expect that the information processed by the encoder can be compressed in latent spaces \citep{razavi2019generating, 8969272}. 

\subsubsection{Feature encoding}
\label{sec:encoder_network}

The encoder uses a two-dimensional (2D) medical image $\bm{x} \in \mathbb{R}^{C \times H \times W}$ as an input, where $C$ is the number of channels and $H$ and $W$ represent the height and width of the images, respectively. Then, the encoder maps the image into two latent representations, $\bm{z}_e^- \in \mathbb{R}^{D \times H^{\prime} \times W^{\prime}}$ and $\bm{z}_e^+ \in \mathbb{R}^{D \times H^{\prime} \times W^{\prime}}$, where $\bm{z}_e^-$ and $\bm{z}_e^+$ correspond to the semantic features of normal and abnormal anatomies, respectively. We use $\bm{z}^\mp_e$ to represent both features. Subsequently, vector quantization is used to discretize $\bm{z}^\mp_e$. Namely, each elemental vector $z^\mp_{e_i} \in \mathbb{R}^{D}$ is replaced with the closest code vector in each codebook $\bm{e}^\mp \in \mathbb{R}^{D \times K}$ that comprises $D$-dimensional $K$ code vectors. Details in this vector quantization process are presented in the next subsection. We denote the quantized vector of $\bm{z}_e^\mp$ as $\bm{z}_q^\mp$. Hereinafter, $\bm{z}_q^-$ is referred to as \emph{normal anatomy code} and $\bm{z}_q^+$ as \emph{abnormal anatomy code}.

\subsubsection{Vector quantization}
\label{sec:vector_quantiser}

The latent space has two codebooks, $\bm{e}^- = \{e^-_k |k = 1, \ldots, K\}\in \mathbb{R}^{K \times D}$ and $\bm{e}^+ = \{e^+_k |k = 1, \ldots, K\}\in \mathbb{R}^{K \times D}$, corresponding to the normal and abnormal semantic features, respectively. The vector quantization process is similar to that of a vector-quantized (VQ) VAEs \citep{oord2017neural}. An $i$-th elemental vector of $\bm{z}^\mp_e$, denoted as $z^\mp_{e_i} \in \mathbb{R}^{D}$, is quantized by executing a nearest-neighbor lookup on the codebook as follows:
\begin{equation} 
k^\mp = \argmin_{k \in \{1, \ldots,  K\}} \|z^\mp_{e_i} - e_k\|_2^2.
\end{equation} 
Thereafter, an $i$-th elemental vector of $\bm{z}^\mp_e$ is replaced by the $k^\mp$-th code vector in the codebook as follows: 
\begin{equation} 
z^\mp_{e_i} = e_{k^\mp}.
\end{equation} 
This replacement is performed for all ($H^\prime \times W^\prime$) elemental vectors of $\bm{z}^\mp_e$ to collectively form a quantized vector $\bm{z}^\mp_q$.

To optimize this process, the encoder and codebooks are updated to minimize an objective, which is referred to as latent loss $L_{\mathrm{lat}}$ as follows:
\begin{equation} 
L^\mp_\mathrm{lat} = \|\mathrm{sg}[\bm{z}^\mp_e] - \bm{e}^\mp\|^2_2 + \beta \|\bm{z}^\mp_e - \mathrm{sg}[\bm{e}^\mp]\|^2_2,
\end{equation} 
\begin{equation} 
L_\mathrm{lat} = L_\mathrm{lat}^- + L_\mathrm{lat}^+,
\end{equation} 
where $\mathrm{sg}$ indicates a stop-gradient operator, which serves as an identity function at the forward computation time and has zero partial derivatives, and $\beta$ is a balancing hyperparameter. During training, the first term in the abovementioned equation updates the codebook variables by delivering the code vectors to the encoder output. Meanwhile, the second term encourages the encoder output to move closer to the targeted code vectors. We use the exponential moving average to train the codebook \citep{kaiser2018fast}.

\subsubsection{Feature decoding}
\label{sec:feature_decoding}

\begin{figure}[t]
  \centering
  \includegraphics[]{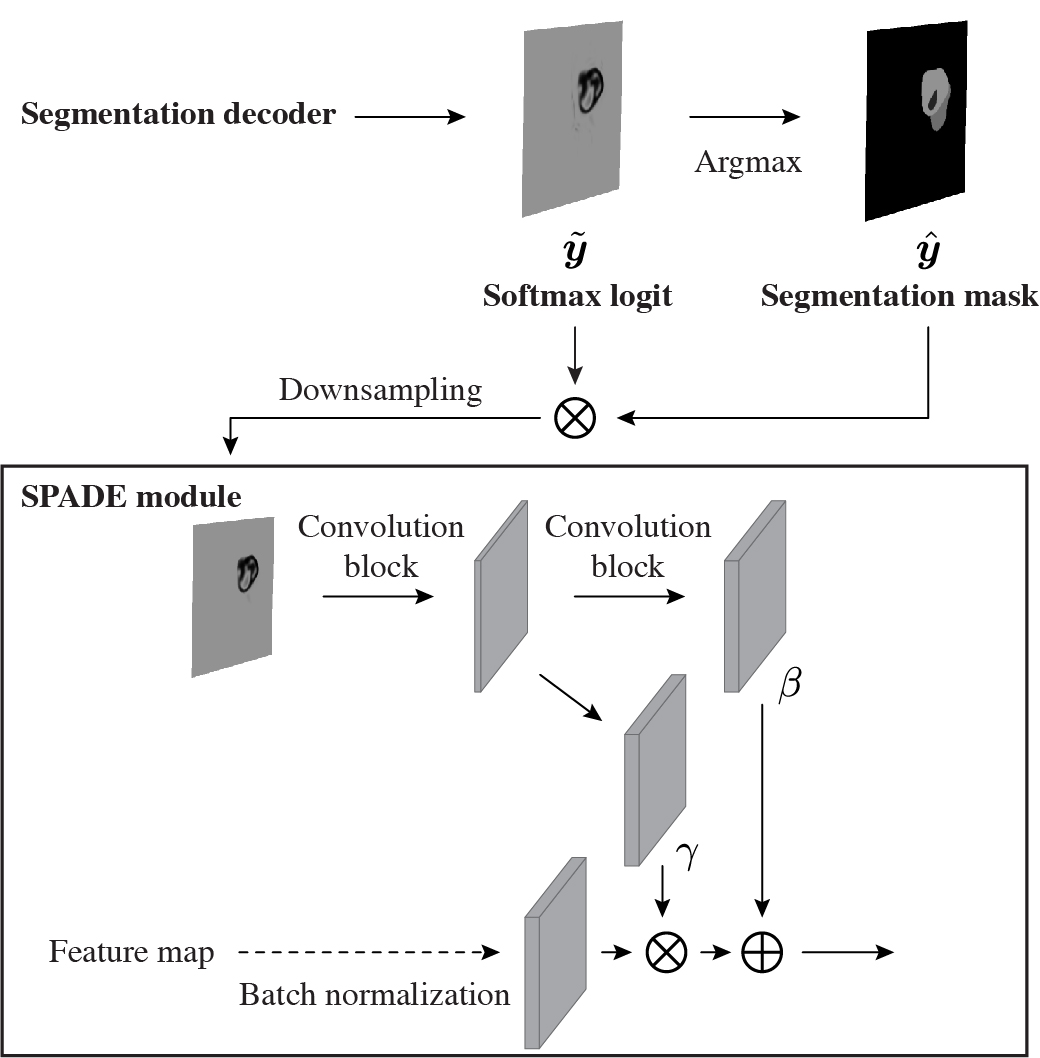}
  \caption{\textbf{Overview of spatially adaptive normalization (SPADE) module.} Softmax logit $\tilde{\bm{y}}$ multiplied by the segmentation mask $\hat{\bm{y}}$ is further downsampled to achieve resolutions corresponding to those of each layer in the image decoder. SPADE module propagates the semantic layout of abnormalities into the image generation process. Each convolution block comprises a convolution function and bias parameter.}
  \label{fig:spade}
\end{figure}

The segmentation decoder uses an abnormal anatomy code $\bm{z}_q^+$ as an input and outputs a $S$-class segmentation label $\hat{\bm{y}} \in \mathbb{R}^{S \times H \times W}$ that corresponds to the ground-truth label $\bm{y}$. A loss function for the segmentation output $L_{\mathrm{seg}}$ is a composite of the generalized Dice \citep{Sudre_2017} and focal \citep{focal2017} losses as follows:
\begin{equation} 
L_\mathrm{dice} = 1 - 2 \frac{\sum_{s \in S} w_s |\hat{\bm{y}}_{s} \cap \bm{y}_{s}|}{\sum_{s \in S} w_s (|\hat{\bm{y}}_{s}| + |\bm{y}_{s}|)},
\end{equation} 
\begin{equation} 
L_\mathrm{focal} = - \frac{1}{N} \sum_{s \in S} \bm{y}_s (1 - \tilde{\bm{y}}_s)^{\gamma} \log \tilde{\bm{y}}_s,
\end{equation} 
\begin{equation} 
L_\mathrm{seg} = L_\mathrm{dice} + L_\mathrm{focal},
\end{equation} 
where $\tilde{\bm{y}}$ indicates the softmax logits of the segmentation decoder, $N (= H \times W)$ is the number of pixels, and $w_k$ is determined as $w_k = \frac{1}{(\sum_N |\bm{y}_s|)^2}$ to mitigate the class imbalance problem \citep{Sudre_2017}.

Meanwhile, the image decoder $f$ performs conditional image generation using the \emph{spatially adaptive normalization} (SPADE) \citep{park2019SPADE}. SPADE is designed to propagate semantic layouts to the process of synthesizing images ({\bf Fig. \ref{fig:spade}}). The image decoder uses a normal anatomy code $\bm{z}_q^-$ as its primary input. When the image decoder is used to reconstruct the entire input image $\hat{\bm{x}}^+$, the softmax logits of the segmentation decoder $\tilde{\bm{y}}$ is transmitted to each layer of the image decoder via the SPADE modules ($f(\bm{z}_q^-, \tilde{\bm{y}}) = \hat{\bm{x}}^+$). When {\it null} information, where $\tilde{\bm{y}}$ is filled with 0s, is propagated to the SPADE modules, a normal-appearing reconstruction $\hat{\bm{x}}^-$ is generated by the image decoder ($f(\bm{z}_q^-, \bm{0}) = \hat{\bm{x}}^-$).

To enforce different characteristics between the two types of generated images, $\hat{\bm{x}}^-$ and $\hat{\bm{x}}^+$, we apply a pixel-wise reconstruction loss depending on the region of abnormality. Suppose $\bm{M}^+ \in \{0, 1\}^{C \times H \times W}$ defines a mask, indicating that pixels with any abnormality labels are set to 1 and 0 otherwise, and $\bm{M}^- = \bm{1} - \bm{M}^+$ is a complementary set of $\bm{M}^+$. Briefly, $\bm{M}^+$ presents the abnormal anatomy region, and $\bm{M}^-$ indicates the normal anatomy region. Using these masks, image reconstruction loss $L_\mathrm{rec}$ is defined as follows: \begin{equation} 
\begin{split}
L^-_\mathrm{rec} &= \|\bm{M}^- \odot \hat{\bm{x}}^- - \bm{M}^- \odot \bm{x}\|^2_2 \\
                            &+ (1 - \mathrm{SSIM}(\bm{M}^- \odot \hat{\bm{x}}^-, \bm{M}^- \odot \bm{x})),
\end{split}
\end{equation} 
\begin{equation}
\begin{split}
L^+_\mathrm{rec} &= \|\hat{\bm{x}}^+ - \bm{x}\|^2_2 \\
                            &+ (1 - \mathrm{SSIM}(\hat{\bm{x}}^+, \bm{x})) \\
                            &+ \|{\bm{M}^+} \odot \hat{\bm{x}}^+ - {\bm{M}^+} \odot \bm{x}\|^2_2 \\
                            &+ (1 - \mathrm{SSIM}({\bm{M}^+} \odot \hat{\bm{x}}^+, {\bm{M}^+} \odot \bm{x})),
\end{split}
\end{equation} 
\begin{equation} 
L_\mathrm{rec} = L^-_\mathrm{rec} + L^+_\mathrm{rec},
\end{equation} 
where SSIM indicates structural similarity \citep{Wang2004}, which is added to the L2 loss as a constraint owing to its empirical effect to stabilize the image generation process. 

\subsubsection{Distribution matching}
\label{sec:density_matching}

It is quite important to ensure that each decomposed feature, normal anatomy code $\bm{z}_q^-$ and abnormal anatomy code $\bm{z}_q^+$, corresponds to targeted semantic content in the images. For example, when some code vectors of normal anatomy codes convey not only features corresponding to normal anatomies but also those related to abnormal anatomies, the feature decomposition can be ``leaky,'' losing its reliability for downstream tasks. Particularly, this can occur at normal anatomy codes because, when a pathological image is provided, it is fundamentally impossible to obtain a \emph{paired} normal counterpart that can be a ground-truth for the normal-appearing reconstructions $\hat{\bm{x}}^-$. Therefore, we utilize \emph{unpaired} healthy images and employ distribution matching technique to minimize the discrepancy of the distributions between normal anatomy codes from healthy images and those from disease images. We consider this discrepancy as the Wasserstein distance $d_\mathrm{W}$ and minimize it using Wasserstein GAN \citep{pmlr-v70-arjovsky17a} with gradient penalty \citep{NIPS2017_892c3b1c} that has a critic network $g$. 

Here, we consider that the set of input images $\mathcal{X}$ can be divided into a set consisting of healthy images $\mathcal{X}_h$ and a set consisting of diseased images $\mathcal{X}_d$. Suppose the distribution of normal anatomy codes $\mathcal{Z}^-_q$ can also be split into those originating from healthy images $\mathcal{Z}^-_h$ and those originating from diseased images $\mathcal{Z}^-_d$. When a batch $\{\bm{x} \sim \mathcal{X}\}$ containing both healthy images $\{\bm{x}_h \sim \mathcal{X}_h\}$ and diseased images $\{\bm{x}_d \sim \mathcal{X}_d\}$ is fed into the encoder, a corresponding batch of normal anatomy codes $\{\bm{z}_q^- \sim \mathcal{Z}_q^- \}$ can be split into those originating from healthy images $\{\bm{z}^-_h \sim \mathcal{Z}^-_h \}$ and diseased images $\{\bm{z}^-_d \sim \mathcal{Z}^-_d \}$. If there is no leakage of semantic content of abnormality into the normal anatomy codes, the two distributions, $\mathcal{Z}^-_h$ and $\mathcal{Z}^-_d$, should be identical with each other, irrespective of the presence of abnormality in the input images. Then, distribution matching imposes two types of loss functions, $L_\mathrm{critic}$ and $L_\mathrm{reg}$, as follows:
\begin{equation} 
L_\mathrm{critic} = d_\mathrm{W} + \lambda_\mathrm{gp} (\| \nabla_{\bm{z}^-_m} g (\bm{z}^-_m) \|^2_2 - 1)^2,
\end{equation} 
\begin{equation} 
L_\mathrm{reg} = - g (\bm{z}^-_d),
\end{equation} 
where $d_\mathrm{W} = g (\bm{z}^-_d) - g (\bm{z}^-_h)$ is the Wasserstein distance, $\lambda_\mathrm{gp}$ is the balancing term for the gradient penalty, and $\bm{z}^-_m = \epsilon \bm{z}^-_h + (1 - \epsilon) \bm{z}^-_d$ is to enforce the Lipschitz constraint by sampling a variable along straight lines between pairs of points sampled from the two latent distributions \citep{NIPS2017_892c3b1c}. When optimizing $L_\mathrm{critic}$, only the critic network is trained, not propagating gradients to the modules prior to $\bm{z}_h^-$ and $\bm{z}_d^-$. Both the encoder and codebook for normal anatomy codes are used to bring the two distributions closer together in the process of optimizing $L_\mathrm{reg}$.

Note that, unlike usual GANs for image synthesis, our goal is to achieve an alignment between two latent distributions, $\mathcal{Z}^-_h$ and $\mathcal{Z}^-_d$. Since distribution matching is applied to the batch containing quantized latent codes, it is expected that certain constraints will be added to the process of vector quantization. More specifically, it can be expected to regularize code vectors in the codebook for normal anatomy codes not to be too representative even for abnormal imaging features. 

\subsubsection{Overall learning objectives}
\label{sec:overall_learning objectives}

In summary, we define several loss functions: latent loss $L_\mathrm{lat}$ for optimization of the encoder and codebooks; segmentation loss $L_\mathrm{seg}$ for the segmentation decoder, encoder, and codebook for abnormal anatomy codes; reconstruction loss $L_\mathrm{rec}$ for the image decoder, encoder, and codebook for normal anatomy codes; and distribution matching loss for the critic network $L_\mathrm{critic}$ and that for the encoder and codebook for normal anatomy codes $L_\mathrm{reg}$. The overall learning algorithm is shown in {\bf Algorithm \ref{alg:learning_algorithm}}. 

\begin{algorithm}[t]
\SetAlgoLined
 $e$: encoder\\
 $v^-$: vector quantiser for normal anatomy code\\
 $v^+$: vector quantiser for abnormal anatomy code\\
 $s$: segmentation decoder\\
 $f$: image decoder\\
 $g$: critic network\\
 $\mathcal{D}$: training dataset\\
 sg: stop-gradient operator\\
 $m$: the number of inner iteration for the critic network\\
 $\lambda_1, \lambda_2, \lambda_3, \lambda_4, \lambda_5$: balancing terms for each loss function\\
 \While{not converge}{
  Sample a batch of images $\bm{x}$ and segmentation labels for abnormal regions $\bm{y}$ from $\mathcal{D}$.\\
  $\bm{z}_e^-, \bm{z}_e^+ \leftarrow e(\bm{x})$\\
  $\bm{z}_q^- \leftarrow v^- (\bm{z}_e^-)$\\
  $\bm{z}_q^+ \leftarrow v^+ (\bm{z}_e^+)$\\
  $\hat{\bm{y}}, \tilde{\bm{y}} \leftarrow s (\bm{z}_q^+)$\\
  $\hat{\bm{x}}^+ \leftarrow f (\bm{z}_q^-, \mathrm{sg}(\tilde{\bm{y}}))$\\
  $\hat{\bm{x}}^- \leftarrow f (\bm{z}_q^-, \bm{0})$\\
  Compute $L_\mathrm{lat}(\bm{z}_e^-, \bm{z}_e^+)$, $L_\mathrm{seg} (\hat{\bm{y}}, \bm{y})$, and $L_\mathrm{rec} (\hat{\bm{x}}^-, \hat{\bm{x}}^+, \bm{x})$.\\
  Split the batch of $\{\bm{z}_q^-\}$ into subbatches of $\{\bm{z}^-_h\}$ and $\{\bm{z}^-_d\}$ according to the presence of abnormal label in each input image.\\
  \For{i = 1, $\dots$, m}{
    $\bm{z}^-_h \leftarrow \mathrm{sg} (\bm{z}^-_h)$\\
    $\bm{z}^-_d \leftarrow \mathrm{sg} (\bm{z}^-_d)$\\
    Sample a random number $\epsilon \sim \mathbb{U}[0, 1]$.\\
    $\bm{z}^-_m \leftarrow \epsilon \bm{z}^-_h + (1 - \epsilon) \bm{z}^-_d$\\
    Compute $L_\mathrm{critic} (\bm{z}^-_h, \bm{z}^-_d, \bm{z}^-_m)$.\\
    Update parameters of $g$ to minimize $\lambda_5 L_\mathrm{critic}$ using stochastic gradient descent (e.g., Adam).\\
  }
  Compute $L_\mathrm{reg} (\bm{z}^-_d)$.\\
  Update parameters of $e$, $v^\mp$, $s$, and $f$ to minimize $\lambda_1 L_\mathrm{lat} + \lambda_2 L_\mathrm{seg} + \lambda_3 L_\mathrm{rec} + \lambda_4 L_\mathrm{reg}$ using stochastic gradient descent (e.g., Adam). 
 }
 \caption{Training of the Feature Decomposing Network}
 \label{alg:learning_algorithm}
\end{algorithm}

\subsection{Modeling of content-based image retrieval}
\label{sec:modeling_of_content-based_image_retrieval}

Here, we formulate the problem of CBIR to find the closest feature vector from a reference database containing $N$ $D$-dimensional database vectors $\{ r_n \}^N_{n = 1}$, given a query vector $q$, as follows:
\begin{equation} 
\argmin_{n \in {1, \ldots, N}} D(q, r_n),
\label{eq:separating_hyperplane}
\end{equation} 
where $D$ is a distance function such as Euclidean distance. 

Starting with the abovementioned equation, our consideration into the CBIR system is twofold. First, to enhance computational efficiency in the distance calculation, we propose to binarize latent representations by leveraging the latent spaces that are constructed as to be discrete rather than continuous. Then, a novel CBIR framework by utilizing the decomposed latent codes to retrieve images with targeted semantic components is introduced. 

\subsubsection{Binary hashing based on separating hyperplanes}
\label{sec:binary_hashing_based_on_separating_hyperplanes}

Here, we consider how to binarize code vectors in a codebook $\bm{e} = \{e_k |k = 1, \ldots, K\}\in \mathbb{R}^{K \times D}$, which is learned in the feature decomposing network. The goal is to find subspaces that can be transversed by Hamming distance rearrangement. Since each code vector $e_k \in \mathbb{R}^D$ has a fixed position in the latent space, the latent space can be divided by $K \choose 2$ separating hyperplanes that perpendicularly bisect a line segment connecting any two code vectors. Given two code vectors, $e_i$ and $e_j$, points $x$ located on the separating hyperplane can be formulated as follows:
\begin{equation} 
H_{(i, j)} (x) = (e_i - e_j) x - \frac{1}{2} (\|e_i\|^2_2 - \|e_j\|^2_2) = 0.
\label{eq:separating_hyperplane}
\end{equation} 
Therefore, the position of another code vector $e_k$ can be binarized according to the side of the separating hyperplanes it is on:
\begin{equation} 
\mathrm{sgn} [H_{i, j} (e_k)], 
\label{eq:separating_hyperplane}
\end{equation} 
where
\begin{equation} 
\mathrm{sgn} (x) = \begin{cases}
  1 & (x \geq 0) \\
  -1 & (x < 0) \\
\end{cases}.
\end{equation} 
Then, by considering the positional relationship between a total of $K \choose 2$ separating hyperplanes, the continuous codebook $\bm{e} \in \mathbb{R}^{K \times D}$ can be converted into a binarized codebook $\bm{b} = \{b_k | k = 1, \dots, K\} \in \mathbb{R}^{K \times E}$, where $E = $ $K \choose 2$. Experimentally, we confirmed that there was no code vector that is exactly located on any separating hyperplane, which would make Eq. (\ref{eq:separating_hyperplane}) equal to zero. Since the binary representation corresponding to the location of each vector is uniquely determined, it can be regarded as binary hashing.

\subsubsection{Optimization of the binarized vector length}
\label{sec:optimization_of_binarized_vector_length}

When naively performing the abovementioned binarization, the length of each binarized code vector is ${K \choose 2} = {512 \choose 2} = 130,816$, which is extremely long to obtain the computational benefit from the Hamming distance calculation. Therefore, we consider optimization for the length of the binarized code vectors. At the time of performing nearest neighbor search, a local sensitivity is primarily required; that is, the relationship between the closest code vectors should remain the same. Meanwhile, it does not necessarily guarantee the positional relationship between distance code vectors. Based on these ideas, we can optimize the length of the binarized code vector by removing each element one by one to investigate whether the proximity between the closest vectors changes or not. This optimization algorithm is shown in {\bf \ref{app:optimization_algorithm}}. Empirically, the length of the code vectors can be reduced to $<$ 1\%, allowing the calculation of Hamming distance to be much faster than that of L2 distance using continuous code vectors. Hereinafter, we denote $\bm{b}^* = \{b_k^* | k = 1, \dots, K\} \in \mathbb{R}^{K \times E^*}$ as a binarized codebook after optimization. 

\begin{figure}[t]
  \centering
  \includegraphics[]{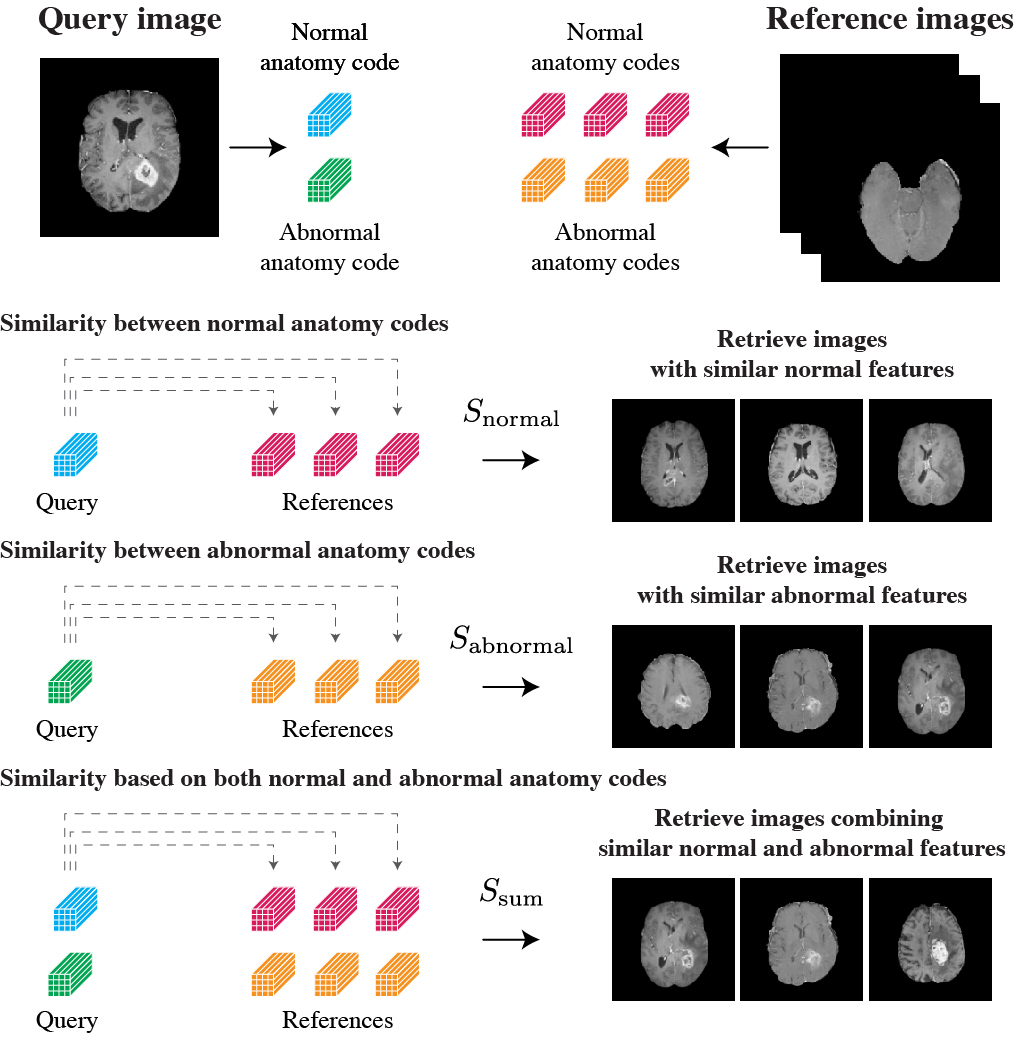}
  \caption{\textbf{Overview of the content-based image retrieval framework.} First, a query image and reference images are decomposed into normal and abnormal anatomy codes. Then, by calculating the similarity of normal anatomy codes between the query image and reference images ($S_\mathrm{normal}$), similar images when viewed as normal anatomies without any abnormality can be retrieved. In contrast, by calculating the similarity of abnormal anatomy codes, images with similar tumor regions can be retrieved ($S_\mathrm{abnormal}$). Besides, the similarity retrieval based on the whole imaging features can be applied by calculating the similarity combining both normal and abnormal anatomy codes ($S_\mathrm{sum}$). Note that these similarity measurements can be calculated based on different distance definitions, such as Euclidean distance, angular distance, and Hamming distance.} 
  \label{fig:cbir_overview}
\end{figure}

\subsubsection{Image retrieval based on the decomposed latent codes}
\label{sec:query_by_image}

When performing the CBIR framework, a query image is decomposed into a normal anatomy code and abnormal anatomy code through the feature decomposing network ({\bf Fig. \ref{fig:cbir_overview}}). Then, the similarity using either the latent codes or a combination of both is calculated between the query image and images in the reference dataset. Particularly, when the similarity is viewed as counterfactual normal images as they should be, the similarity measurement uses only normal anatomy codes to be compared $S_\mathrm{normal} (q, r)$. In contrast, when viewed by focusing only on abnormal areas, the similarity measurement uses only abnormal anatomy codes $S_\mathrm{abnormal} (q, r)$. Moreover, to calculate similarities of the whole imaging features between the query and reference images, we define $S_\mathrm{sum} (q, r)$ for the summation of both measurements as follows:
\begin{equation} 
\label{eq:dsum}
S_\mathrm{sum} (q, r) = S_\mathrm{normal} (q, r) + S_\mathrm{abnormal} (q, r).
\end{equation} 

These similarity measurements can be calculated based on different distance definitions. Here, for comparison, three types of distances, that is, Euclidean distance $D_\mathrm{E}$, angular distance $D_\mathrm{A}$, and Hamming distance $D_\mathrm{H}$, are calculated for each similarity measurement. Note that the continuous codebook $\bm{e}$ is the basis for Euclidean distance $D_\mathrm{E}$ and angular distance $D_\mathrm{A}$, while the optimized binarized codebook $\bm{b}^*$ is the basis for Hamming distance $D_\mathrm{H}$.

\section{Experiments}
\label{sec:experiments}

\subsection{Dataset}
\label{sec:dataset}

We used brain MR images of gliomas from the 2019 BraTS Challenge \citep{6975210brats, Bakas2017, TCGAGBM, TCGALGG}, containing a dataset of 355 patients for training (MICCAI\_BraTS\_Training), a dataset of 125 patients for validation (MICCAI\_BraTS\_Validation), and a dataset of 167 patients (MICCAI\_BraTS\_Testing) for testing. For each patient, T1, Gd-enhanced T1, T2, and fluid-attenuated inversion recovery (FLAIR) sequences were obtained. MICCAI\_BraTS\_Training contains three segmentation labels of abnormality: Gd-enhancing tumor (ET), peritumoral edema (ED), and necrotic and non-enhancing tumor core (NET). Conversely, MICCAI\_BraTS\_Validation and MICCAI\_BraTS\_Testing do not have any segmentation label. There is no ground-truth that represents normal anatomical structures. Therefore, we segmented all three datasets into six normal anatomical labels (left cerebrum, right cerebrum, left cerebellum, right cerebellum, left ventricle, and right ventricle). Moreover, the abnormal regions (ET, ED, and NET) in MICCAI\_BraTS\_Validation and MICCAI\_BraTS\_Testing were segmented in the same manner with those in MICCAI\_BraTS\_Training. The annotation process is described in detail in {\bf \ref{app:annotation_process}}.

For the training of the feature decomposing network, we concatenated both MICCAI\_BraTS\_Validation and MICCAI\_BraTS\_Testing as a \emph{training dataset}. Then, the feature decomposing network was evaluated on MICCAI\_BraTS\_Training as a \emph{test dataset}, which is also utilized in the demonstration of the CBIR system based on the decomposed latent codes. 

\subsection{Preprocessing}
\label{sec:pre-processing}

All four sequences, T1, Gd-enhanced T1, T2, and FLAIR, were concatenated into a four-channel MR volume $\bm{X} \in \mathbb{R}^{4 \times 240 \times 240 \times 155}$. Then, a preprocessing pipeline, including axial image resizing to $256 \times 256$ and Z-score normalization, was performed. Subsequently, each three-dimensional (3D) MR volume was decomposed into a collection of 2D axial slices $\{\bm{x}_1, \bm{x}_2, \dots, \bm{x}_{155} \in \mathbb{R}^{4 \times 256 \times 256} \}$. The training and test datasets underwent the same preprocessing process. Data augmentation, such as random rotation and random horizontal flip, was applied to each image in the training dataset to train the feature decomposing network.

\subsection{Implementation}
\label{sec:implementation}

All experiments were implemented in Python 3.7 with PyTorch library 1.2.0 \citep{NEURIPS2019_9015} using an NVIDIA Tesla V100 graphics processing unit and CUDA 10.0. For all networks, Adam optimization \citep{kingma2014adam} was used for the training. Network initialization was performed using the method described by He et al. \citep{he2015delving}.

\subsubsection{Feature decomposing network}
\label{sec:implement_feature_decomposing_network}

When using a quantized latent space, such as VQ-VAE, the size of latent representation (i.e., width and height of the feature maps) exerts a significant effect on the quality of image generation \citep{razavi2019generating}. Therefore, for comparison, we configured several architectures of the feature decomposing network depending on the size of the latent space. Hereinafter, we denote a feature decomposing network with the latent representation of the size of $H^\prime \times W^\prime$ as $\mathrm{FDN}_{H^\prime \times W^\prime}$. In this study, we compared $\mathrm{FDN}_{4 \times 4}$, $\mathrm{FDN}_{8 \times 8}$, $\mathrm{FDN}_{16 \times 16}$, and $\mathrm{FDN}_{32 \times 32}$ with or without the distribution matching for the normal anatomy codes. 

As described in \textbf{Section \ref{sec:training_of_feature_decomposing_network}}, the feature decomposing network comprises the encoder, segmentation decoder, and image decoder. See {\bf \ref{app:detailed_architecture}} for the detail of the architectures of these neural networks. The input for the encoder is required to be a four-channel 2D image with the size of $4 \times 256 \times 256$ $(= \mathrm{channel} \times \mathrm{height} \times \mathrm{width})$. The encoder has a common trunk for obtaining an input image and extracting low-level features of the image and then bifurcates into two branches with the same architecture, one of which is for the normal anatomy code and the other is for the abnormal anatomy code. The number of a repeated structure, consisting of residual blocks \citep{He2015} with a strided convolution (ResBlock + StridedConv), was adaptively set for each size of the latent representation. For example, in $\mathrm{FDN}_{8 \times 8}$, the encoder utilized $32-64-128-128-128-128$ filter kernels in each layer. The encoder outputs two latent representations corresponding to normal and abnormal semantic components, $\bm{z}_e^-$ and $\bm{z}_e^+$, with the size of $64 \times H^\prime \times W^\prime$. These latent representations are subsequently quantized to $\bm{z}_q^-$ and $\bm{z}_q^+$ through the vector quantization. The dimension of code vectors $D$ and number of code vectors $K$ in the codebooks were fixed as follows: $D = 64$, and $K = 512$.

The image and segmentation decoders have almost the same architecture, except for the normalization layer, where the image decoder particularly utilizes SPADE module for the propagation of softmax logits $\tilde{\bm{y}}$ from the segmentation network. The number of a repeated structure, which comprises upsampling module with a bilinear interpolation function followed by residual block with or without SPADE [Upsample + (SPADE-)ResBlock], was adopted for the size of the latent representation. For example, in $\mathrm{FDN}_{8 \times 8}$, the image and segmentation decoders utilized $128-128-128-128-64-32$ filter kernels in each layer. The segmentation decoder takes $\bm{z}_q^+$ as an input and outputs a segmentation map $\hat{\bm{y}}$ with the size of $4 \times 256 \times 256$ $(= \mathrm{channel} \times \mathrm{height} \times \mathrm{width})$, the channel number of which corresponds to the number of abnormality labels (ET, ED, and TC) plus a background label. Besides, a softmax logit $\tilde{\bm{y}}$ with the size of $4 \times 256 \times 256$ is retained for the collateral input for the image decoder. The image decoder utilizes $\bm{z}_q^-$ as primary input. Depending on the presence or absence of the softmax logit $\tilde{\bm{y}}$ through the SPADE modules, it generates either a normal-appearing reconstruction $\hat{\bm{x}}^-$ or entire reconstruction of the input image $\hat{\bm{x}}^+$, respectively. 

When using distribution matching to regularize normal anatomy codes (see \textbf{Section \ref{sec:density_matching}}), the critic network is trained to approximate the Wasserstein distance $d_\mathrm{W}$ of distributions between the normal anatomy codes originating from healthy images $\mathcal{Z}_h^-$ and those originating from diseased images $\mathcal{Z}_d^-$. The detailed network architecture is described in {\bf \ref{app:detailed_architecture}}. The number of inner iterations in the training of the critic network $m$ was set to 5, and the balancing term for the gradient penalty $\lambda_{\mathrm{gp}}$ was 10.0. To balance the magnitudes of the loss functions, $\lambda_4$ and $\lambda_5$ were set to $1.0 \times 10^{-4}$ for $\mathrm{FDN}_{4 \times 4}$ and $\mathrm{FDN}_{8 \times 8}$ and $5.0 \times 10^{-4}$ for $\mathrm{FDN}_{16 \times 16}$ and $\mathrm{FDN}_{32 \times 32}$. The larger values of $\lambda_4$ and $\lambda_5$ for each configuration tended to fail in mode collapse, where the encoder learned only a few modes of data in the sample distribution. The other hyperparameters were shared across the configurations as follows: batch size = 112, number of training epochs = 400, learning late = $1.0 \times 10^{-4}$, weight decay = $1.0 \times 10^{-5}$, $\lambda_1 = 0.25$, $\lambda_2 = 5.0$, $\lambda_3 = 5.0$, and $\gamma = 2.0$.

\subsubsection{Preparation of codebooks}
\label{sec:content-based_image_retrieval_system}

The CBIR framework was constructed on a per-image basis; that is, each MR volume was separated into slices along the axial axis, which was in the same manner with the training of the feature decomposing network. By straightforwardly applying the trained feature decomposing network, every single image in the test dataset was decomposed into normal anatomy code $\bm{z}_q^-$ and abnormal anatomy code $\bm{z}_q^+$. These latent codes can be considered as a set of code vectors extracted from the continuous codebooks $\bm{e}^\mp$, which is subjective to the calculation of Euclidean distance $D_\mathrm{E}$ and angular distance $D_\mathrm{A}$. Then, according to the methods described in \textbf{Section \ref{sec:binary_hashing_based_on_separating_hyperplanes}} and \textbf{Section \ref{sec:optimization_of_binarized_vector_length}}, we prepared a set of latent codes in a manner of binarized code vectors. First, code vectors with L2 norm $<$ $1.0 \times 10^{-5}$ were rounded to a zero vector. Then, each code vector was binarized with optimized length to obtain the optimized binarized codebooks $\bm{b}^{\mp*}$. Binarized latent codes were acquired by searching the optimized binarized codebooks based on the same indices of the corresponding latent codes in the continuous codebooks. The similarity between binarized code vectors was evaluated based on Hamming distance $D_\mathrm{H}$. 

\subsection{Evaluation}
\label{sec:evaluation}

Since end-to-end learning of the whole modules cannot be performed, we evaluated individual component to find the optimal combination for the CBIR framework. The evaluation process comprised three stages. First, to find the optimal configuration of the feature decomposing network, we assessed the error of image reconstruction, performance of abnormality segmentation, ``leakiness'' of feature decomposition, and compactness of the codebooks. Second, based on the feature decomposing network with a selected configuration, we observed the extent to which the distance relationship between code vectors was changed through the binarization. Lastly, we demonstrated the effectiveness of the proposed CBIR framework based on the decomposed latent codes from qualitative and quantitative perspectives.

\subsubsection{Image reconstruction error}
\label{sec:image_reconstruction_error}

The image decoder performs conditional image generation while switching the entire reconstructions $\hat{\bm{x}}^+$ and normal-appearing reconstructions $\hat{\bm{x}}^-$ (see \textbf{Section \ref{sec:feature_decoding}}). At the training stage, the entire reconstructions are similar to the input images. Meanwhile, normal-appearing reconstructions learn to match the region of the input images that excludes the abnormal area, which can be indicated by the mask matrix $\bm{M}^-$. Therefore, the reconstruction error of the entire reconstructions and that of normal-appearing reconstructions can be calculated as $\sum \| \bm{x} - \hat{\bm{x}}^+ \|_1$ and $\sum \| \bm{M}^- \odot \bm{x} - \bm{M}^- \odot \hat{\bm{x}}^- \|_1$, respectively, where $\sum$ denotes pixel-wise summation of the residual errors.

\subsubsection{Segmentation performance}
\label{sec:segmentation_performance}

The segmentation decoder predicts a segmentation output $\hat{\bm{y}}$, which should be close to the ground-truth label $\bm{y}$ (see \textbf{Section \ref{sec:feature_decoding}}). The segmentation performance was evaluated based on Dice score \citep{dice1945} with respect to the abnormality labels (ET, ED, and NET). The segmentation outputs in each 2D axial slice $\{\bm{x}_1, \bm{x}_2, \dots, \bm{x}_{155}\}$ were concatenated into a volume to evaluate the Dice score based on the volume. 

\subsubsection{Leakiness of feature decomposition}
\label{sec:leakiness_of_feature_decomposition}

As described in \textbf{Section \ref{sec:density_matching}}, distribution matching is introduced to ensure that each decomposed latent code corresponds to targeted semantic content of the input images without being ``leaky'' to other features. Especially, this is important for normal anatomy codes because there is no ground-truth for normal-appearing reconstructions originating from diseased images. However, observation of the latent space cannot provide information on whether the representations contained therein are decomposed in a desirable manner. Hence, we observe reconstructed images generated from these latent codes through the image decoder. In the evaluation, a batch containing only the normal-appearing reconstructions $\{\hat{\bm{x}}^-\}$ was fed into a classification network to predict whether each normal-appearing reconstruction was derived from healthy images $\mathcal{X}_h$ or diseased images $\mathcal{X}_d$. The architecture of the classification network and training settings based on the training dataset are described in {\bf \ref{app:detailed_architecture_classifier}}. When the normal-appearing reconstructions do not contain any abnormal characteristics, the classification network should not be possible to identify their origin, even if they are derived from diseased images. Otherwise, it indicates that there are ``leaked'' abnormal imaging features in the normal-appearing reconstructions, which can be distinguishable for the origin from the diseased images. Positive predictive value (PPV) of the classification network with respect to the origin from the diseased images was evaluated as an indicator of leakiness in the test dataset. 

\subsubsection{Compactness of the codebooks}
\label{sec:compactness_of_codebooks}

In some configurations of the feature decomposing network, it was empirically observed that a portion of the $K$ code vectors stored in a codebook exhibited small norms close to zero (see \textbf{\ref{app:distribution_of_norm}}). It suggested that these code vectors with small norms did not encode distinguishable features with zero vector, allowing approximation as the zero vector. Therefore, in that case, the number of code vectors that are actually significant was $< K$. Hence, we define the compactness of the codebooks to be the ratio of insignificant code vectors that showed small L2 norms below a threshold values of $1.0 \times 10^{-5}$ to the total number of code vectors $K$ in a codebook. When the value of the compactness is large, it implies that the codebook can be considered as compact with less than initial $K$ code vectors. This is preferential for the computational cost at the time of similarity search because we can approximately reduce the number of code vectors to be computed.

\subsubsection{Validity of the binarization process}
\label{sec:validity_of_binarization_process}

The validity of the binarization process, which is introduced in \textbf{Section \ref{sec:binary_hashing_based_on_separating_hyperplanes}} and \textbf{Section \ref{sec:optimization_of_binarized_vector_length}}, was evaluated from three perspectives: concordance of the distance relationship of the code vectors before and after the binarization process, compression ratio of the code vector length, and comparison of computational time. To demonstrate concordance, we investigated whether the nearest neighbor relationship among binarized code vectors changed with respect to the continuous ones. Since the fundamental element of the latent space composed a fixed set of code vectors, the distance relationship as a whole can be regarded as almost unchanged, given that the distance relationship between each code vector is consistent. Therefore, the distance relationship of each code vector with the other remaining $K - 1$ code vectors was assessed and compared between the two types of codebooks. In the evaluation, for each code vector, the indices of the top-$Q$ closest code vectors were obtained. The distance calculation was performed using Hamming distance $D_\mathrm{H}$ for the optimized binarized codebook $\bm{b}^*$ and Euclidean distance $D_\mathrm{E}$ for the continuous codebook $\bm{e}$. Then, the concordance was calculated with respect to the agreement by the Jaccard similarity coefficient between two sets of indices. The compression ratio of the code vectors owing to the optimization process was calculated as follows: $\frac{E^*}{E}$. See {\bf \ref{app:comparison_of_computational_time}} for the methodological details in the comparison of computational time in the distance calculation.

\subsubsection{CBIR performance based on the decomposed latent codes}
\label{sec:eval_query_by_image}

To quantify the image retrieval performance using the decomposed latent codes (see \textbf{Section \ref{sec:query_by_image}}), images containing the largest area of abnormality were selected from each MR volume in the test dataset based on ground-truth labels. We refer to these representative images as \emph{query images}. For each query image, a \emph{reference dataset} that comprised the rest of the images in the test dataset except for the images in the same MR volume of the query image was constructed. The images in the reference dataset are called \emph{reference images}. Then, for each MR volume in the reference dataset, the image with the closest latent code to that of the query image was obtained. The obtained images from each MR volume were sorted with respect to the similarity. Lastly, top-$Q$ closest images, each of which belonged to different MR volume, were provided as retrieved images. This MR volume-wise retrieval can be appropriate in evaluating variable retrieved images by a single query image because one MR volume usually contains several images with similar appearance in continuous axial slices. 

To report retrieval performance, the mean Dice coefficient based on two types of ground-truth labels (see \textbf{Section \ref{sec:dataset}}), one is for six categories of normal anatomy and the other is for three categories of abnormality, was assessed between each query image and top-$Q$ closest images from the reference dataset. The mean Dice coefficient was calculated for all query images and then averaged. As shown in \textbf{Fig. \ref{fig:cbir_overview}}, when using $S_\mathrm{normal}$, the mean Dice coefficients based on the categories of normal anatomy should be high because the similarity is evaluated based on the normal anatomy codes that correspond to normal-appearing reconstructions. Conversely, image retrieval using $S_\mathrm{abnormal}$ should be accompanied by high mean Dice coefficients in the categories of the abnormalities because it evaluates the similarity based on the abnormal anatomy codes relevant to tumor segmentation labels. Therefore, the mean Dice coefficient based on the ground-truth labels of the normal or abnormal anatomical categories was reported for the image retrieval using $S_\mathrm{normal}$ or $S_\mathrm{abnormal}$, respectively. When it comes to the similarity retrieval using $S_\mathrm{sum}$, the mean Dice coefficients based on the ground-truth labels of both normal and the abnormal anatomical categories were averaged because the similarity measurement should represent the features of the whole images. 

For comparison, a \emph{brutal search} to directly maximize Dice overlap was performed for each query image to retrieve top-$Q$ closest images. In comparing the retrieval performance using $S_\mathrm{normal}$ or $S_\mathrm{abnormal}$, the brutal search retrieved images by maximizing Dice coefficients between the ground-truth labels of the normal or abnormal anatomical categories of a query image and those of reference images, respectively. For the similarity measurement using $S_\mathrm{sum}$, the brutal search maximized the simple summation of Dice coefficients of the ground-truth labels of the normal and abnormal anatomical categories. Similar image retrieval using the brutal search was conducted in the same MR volume-wise manner. While the brutal search requires a significant computational time and ground-truth labels for all query and reference images, the mean Dice coefficients obtained can be used as an oracle (technical upper bound). 

\section{Results}
\label{sec:results}

\begin{figure*}[t]
  \centering
  \includegraphics[]{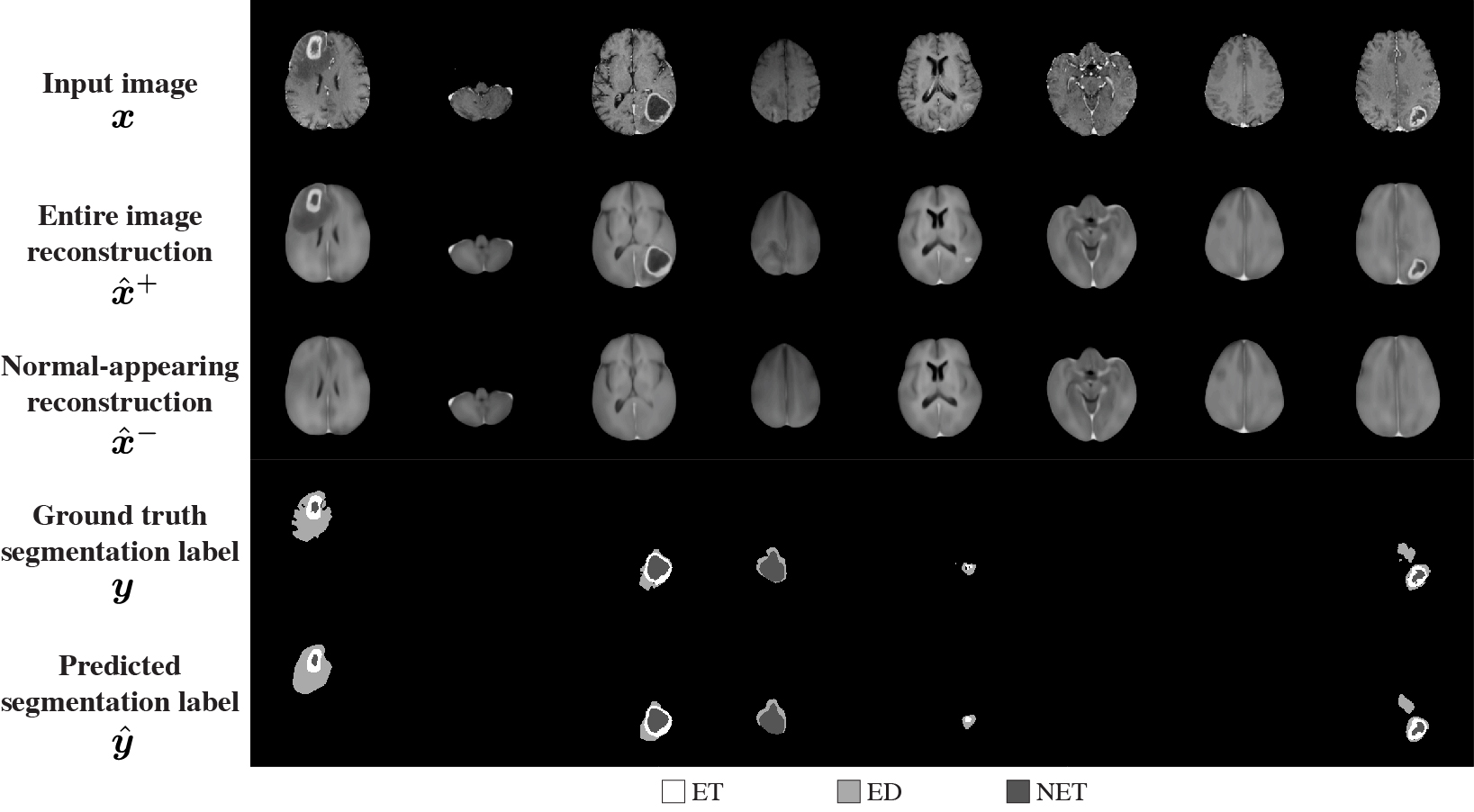}
  \caption{\textbf{Example results of model training based on $\boldsymbol{\mathrm{FDN}_{8 \times 8}}$ with distribution matching.} The first row indicates the input images $\bm{x}$. Entire input images were reconstructed $\hat{\bm{x}}^+$ (second row) based on both normal and abnormal anatomy codes, whereas normal-appearing reconstructions $\hat{\bm{x}}^-$ (third row) were generated on normal anatomy code only. A clear distinction can be observed between $\hat{\bm{x}}^+$ and $\hat{\bm{x}}^-$ at abnormal regions, which existed in both $\bm{x}$ and $\hat{\bm{x}}^+$ but not in $\hat{\bm{x}}^-$. The fourth and fifth rows indicate ground-truth segmentation labels $\bm{y}$ representing the tumor region and prediction for the labels $\hat{\bm{y}}$, respectively. The output of segmentation labels tended to be rounded and did not recover the detailed shape of each region. We consider this as a natural consequence since the compressed representation in the latent codes, which is advantageous for the computational cost of similarity search, did not have sufficient capacity to preserve the detailed feature in the input image as a tradeoff. ET, Gd-enhancing tumor; ED, peritumoral edema; NET, necrotic and non-enhancing tumor core.}
  \label{fig:example_training_results}
\end{figure*}

Here, we first present an example training process of the feature decomposing network. Then, several configurations of the feature decomposing network, particularly focusing on the size of the latent representation and use of distribution matching, are compared to select a model to be exploited for the downstream task. Subsequently, using the feature decomposing network with the selected configuration, we confirm the validity of binary hashing. Lastly, the CBIR system utilizing decomposed latent codes is qualitatively and quantitatively evaluated. 

\subsection{Example training results of the feature decomposing network}
\label{sec:training_results_of_the_feature_decomposing_network}

Among several configurations of the feature decomposing network, an example of the training results of $\mathrm{FDN}_{8 \times 8}$ with distribution matching at epoch 400 is shown in {\bf Fig. \ref{fig:example_training_results}}. The first, second, and third rows demonstrate the input images $\bm{x}$, corresponding entire image reconstructions $\hat{\bm{x}}^+$, and normal-appearing reconstructions $\hat{\bm{x}}^-$, respectively. We can observe that there is a clear distinction between two types of reconstructions, $\hat{\bm{x}}^+$ and $\hat{\bm{x}}^-$, especially when evaluating abnormal regions, which existed in the input images $\bm{x}$ and entire reconstructions $\hat{\bm{x}}^+$ but disappeared in the normal-appearing reconstructions $\hat{\bm{x}}^-$. Note that the regions where the abnormality had existed originally were replaced by imaging characteristics of normal neuroanatomy in the normal-appearing reconstructions $\hat{\bm{x}}^-$. This indicates the complementary capacity derived from normal anatomy codes. Besides, due to the shared normal anatomy codes between the two types of reconstructions, $\hat{\bm{x}}^+$ and $\hat{\bm{x}}^-$, the appearances outside the region of abnormality seems to be almost identical with each other. Furthermore, the fourth and fifth rows in {\bf Fig. \ref{fig:example_training_results}} indicate ground-truth segmentation labels for the abnormality (ET, TC, and NET) and prediction for the labels through the segmentation decoder, respectively. The output of segmentation labels tended to be rounded, without precisely recovering the detailed shape of each tumor region. This can be expected because the compressed representation of the latent space, where the spatial resolution is only $8 \times 8$, is advantageous for computational efficiency at the time of similarity search. For the same reason, we did not pursue the generation quality of the reconstructed images as a primary purpose. Although the detailed part of the textual appearance as realistic MR images was not perfectly reproduced, we consider that it is sufficient for recognizing anatomical location and presence of abnormality in the reconstructed images. The learning process of this example model is demonstrated in {\bf \ref{app:example_learning_process}}.

\subsection{Comparison between several configurations of the feature decomposing network}
\label{sec:comparison_between_several_configuration}

\begin{figure}[p]
  \centering
  \includegraphics[]{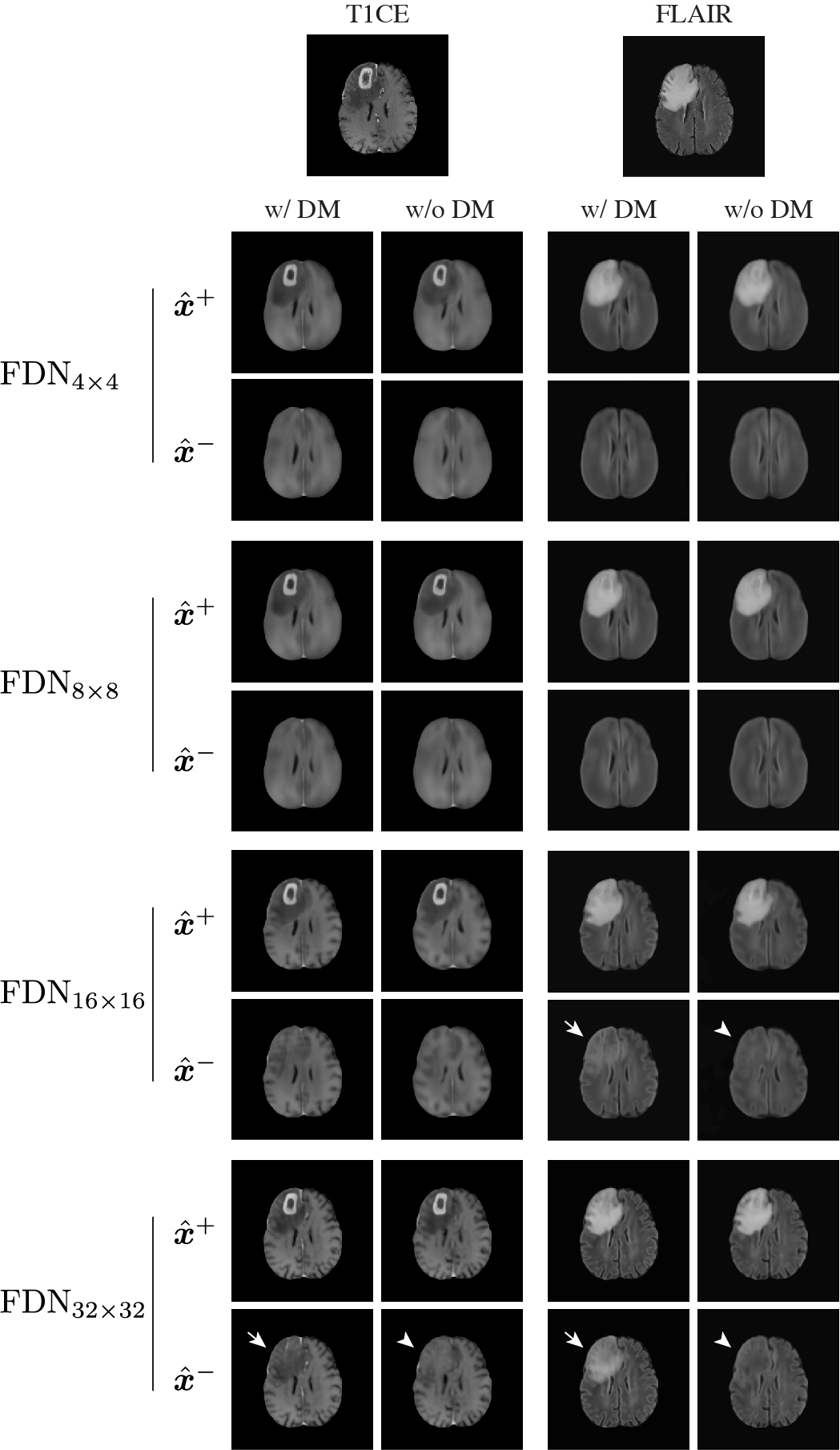}
  \caption{\textbf{Comparison of reconstructed images according to different configurations of the feature decomposing network.} As the spatial resolution of the latent representation increases from $\mathrm{FDN}_{4 \times 4}$ to $\mathrm{FDN}_{32 \times 32}$, visual fidelity of the reconstructed images tends to improve. Note that there are partially reconstructed abnormal regions (arrows) even in the normal-appearing images $\hat{\bm{x}}^-$ through the models with relatively high resolutions (i.e., $\mathrm{FDN}_{16 \times 16}$ and $\mathrm{FDN}_{32 \times 32}$). These ``leaky'' appearances can be alleviated by imposing the distribution matching (arrowheads). DM, distribution matching.}
  \label{fig:comparison_reconstructions}
\end{figure}

To evaluate the effects of the size of the latent representation and use of the distribution matching, we compared several configurations of the feature decomposing network, such as $\mathrm{FDN}_{4 \times 4}$, $\mathrm{FDN}_{8 \times 8}$, $\mathrm{FDN}_{16 \times 16}$, and $\mathrm{FDN}_{32 \times 32}$, with or without distribution matching (see \textbf{Section \ref{sec:implement_feature_decomposing_network}}). See {\bf Fig. \ref{fig:comparison_reconstructions}} for visual results, where two types of reconstructed images, entire image reconstructions $\hat{\bm{x}}^+$ and normal-appearing reconstructions $\hat{\bm{x}}^-$, according to an input image are shown for each configuration. As the resolution of the latent space increased from $4 \times 4$ to $32 \times 32$, the quality of the reconstructed images was improved, showing that fine textures of the brain MR images were reproduced more realistically. As for the difference between entire reconstructions $\hat{\bm{x}}^+$ and normal-appearing reconstructions $\hat{\bm{x}}^-$, it is expected that there should be differences in the areas that correspond to the abnormal sites. Thus, the abnormal areas should appear only in the entire reconstructions $\hat{\bm{x}}^+$ and should be diminished in the normal-appearing reconstructions $\hat{\bm{x}}^-$. When the resolution of the latent representation is relatively low (i.e., $\mathrm{FDN}_{4 \times 4}$ and $\mathrm{FDN}_{8 \times 8}$), the difference was clear even without distribution matching. Conversely, particularly in $\mathrm{FDN}_{16 \times 16}$ and $\mathrm{FDN}_{32 \times 32}$, the abnormal regions, which exhibited as low-intensity area in Gd-enhanced T1 sequence and high-intensity area in FLAIR sequence, were partly reproduced even in the normal-appearing reconstructions $\hat{\bm{x}}^-$ especially when distribution matching was not utilized (see arrows in {\bf Fig. \ref{fig:comparison_reconstructions}}). As mentioned in \textbf{Section \ref{sec:leakiness_of_feature_decomposition}}, we call this failure in decomposing representations as ``leakiness'' of abnormal features into normal anatomy codes $\bm{z}_q^-$. Note that the distribution matching imposed on the normal anatomy codes mitigated this leakiness and encouraged normal-appearing reconstructions to replace the region of abnormality with normal imaging features that would have existed therein if the sample is healthy (indicated by arrowheads in {\bf Fig. \ref{fig:comparison_reconstructions}}). 

\begin{figure}[t!]
  \centering
  \includegraphics[]{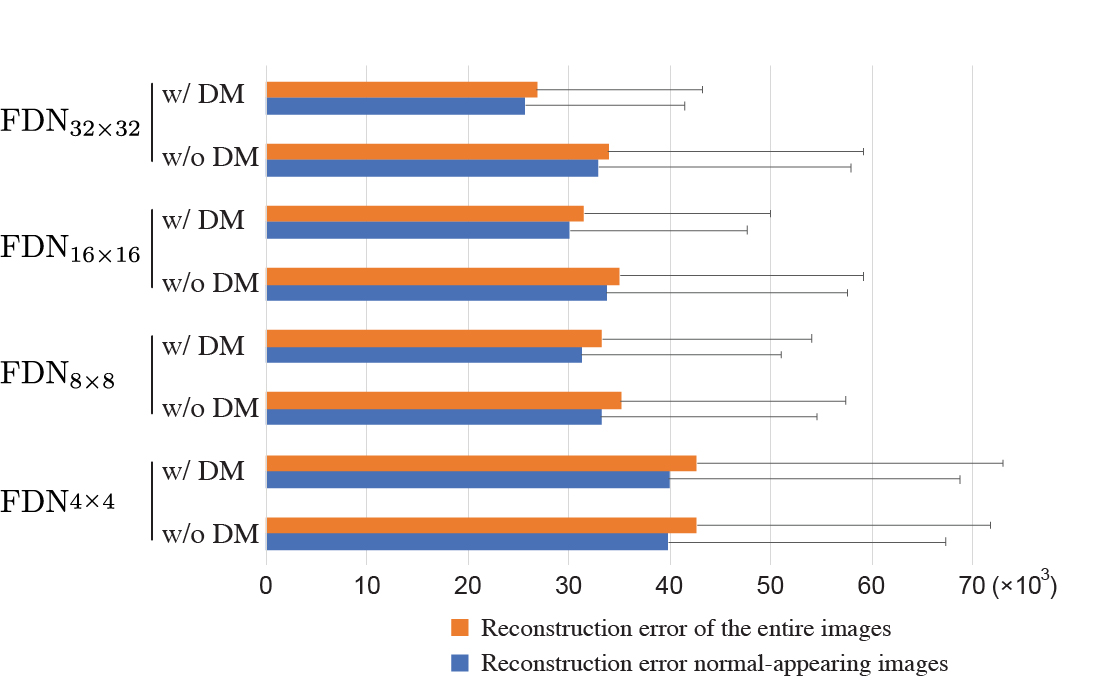}
  \caption{\textbf{Image reconstruction results.} Mean $\pm$ standard deviation of the image reconstruction error (see \textbf{Section \ref{sec:image_reconstruction_error}}) is shown for each configuration of the feature decomposing network. As the resolution of the latent representation increased from $\mathrm{FDN}_{4 \times 4}$ to $\mathrm{FDN}_{32 \times 32}$, the quality of image reconstruction was improved with decreased reconstruction errors. Note that the distribution matching did not cause any negative effect. FDN, feature decomposing network; DM, distribution matching.}
  \label{fig:image_recon_results}
\end{figure}

\begin{figure}[t!]
  \centering
  \includegraphics[]{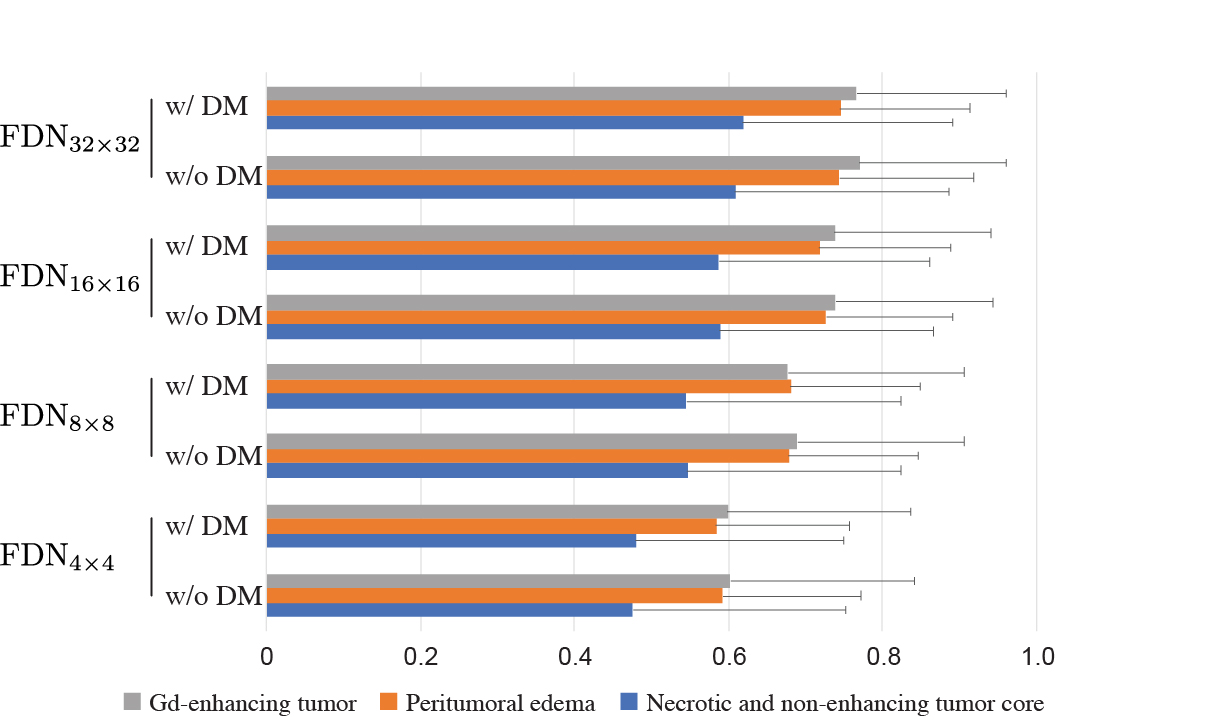}
  \caption{\textbf{Segmentation results.} Mean $\pm$ standard deviation of the averaged Dice scores for the abnormality labels (see \textbf{Section \ref{sec:segmentation_performance}}) is shown for each configuration of the feature decomposing network. As the resolution of the latent representation increased from $\mathrm{FDN}_{4 \times 4}$ to $\mathrm{FDN}_{32 \times 32}$, the segmentation performance was improved. Note that the distribution matching did not cause any negative effect. FDN, feature decomposing network; DM, distribution matching.}
  \label{fig:segmentation_results}
\end{figure}

\begin{figure}[t!]
  \centering
  \includegraphics[]{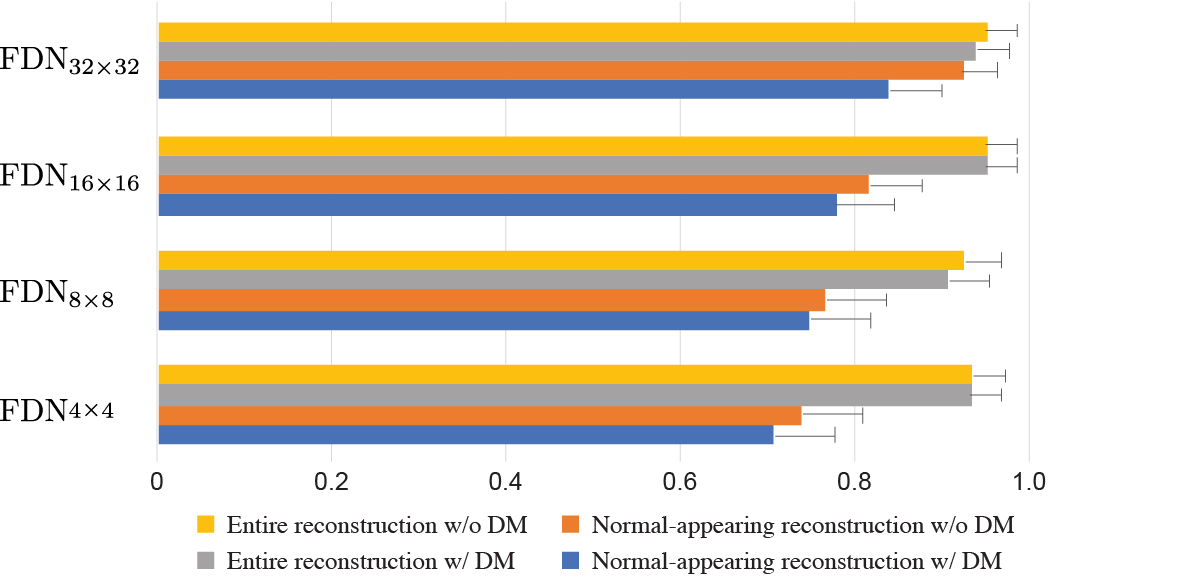}
  \caption{\textbf{Evaluation of the leakiness in the normal-appearing reconstructions.} A classification network was trained based on four different reconstructions, such as entire reconstructions, entire reconstructions with distribution matching, normal-appearing reconstructions, and normal-appearing reconstructions with distribution matching, to classify whether the original images included the abnormality or not (see \textbf{Section \ref{sec:leakiness_of_feature_decomposition}}). Here, leakiness of feature decomposition means the tendency that the normal-appearing images unintentionally contained distinguishable features of abnormality. Mean $\pm$ standard deviation of positive predictive values (PPVs) for the origin from diseased images measured in the test dataset are shown as an indicator of the leakiness. As the latent resolution increased, the PPVs increased (orange bars), especially without the distribution matching. This leakiness of abnormal features was alleviated by imposing the distribution matching (blue bars). FDN, feature decomposing network; DM, distribution matching.}
  \label{fig:leakiness_results}
\end{figure}

\begin{table}[t]
\resizebox{\linewidth}{!}{\begin{tabular}{@{}lccc@{}}
\toprule
\textbf{FDN model}                              & \textbf{Normal anatomy code} & \textbf{Abnormal anatomy code} \\ \midrule
$\mathrm{FDN}_{4 \times 4}$ w/ DM               & 0.79                         & 0.75                           \\
$\mathrm{FDN}_{4 \times 4}$ w/o DM              & 0.75                         & 0.72                           \\
$\mathrm{FDN}_{8 \times 8}$ w/ DM               & 0.68                         & 0.69                           \\
$\mathrm{FDN}_{8 \times 8}$ w/o DM              & 0.63                         & 0.67                           \\
$\mathrm{FDN}_{16 \times 16}$ w/ DM             & 0.71                         & 0.52                           \\
$\mathrm{FDN}_{16 \times 16}$ w/o DM            & 0.25                         & 0.57                           \\
$\mathrm{FDN}_{32 \times 32}$ w/ DM             & 0.31                         & 0.0                            \\
$\mathrm{FDN}_{32 \times 32}$ w/o DM            & 0.0                          & 0.0                            \\ \bottomrule
\end{tabular}}
\caption{\textbf{Compactness of the codebooks.} The compactness, which is the ratio of insignificant code vectors showing small L2 norms below a threshold to the total number of the code vectors (see \textbf{Section \ref{sec:compactness_of_codebooks}}), is shown for each configuration of the feature decomposing network. FDN, feature decomposing network; DM, distribution matching.} 
\label{tab:compactness_of_codebooks}
\end{table}

Subsequently, we quantitatively compared several configurations based on the reconstruction error (see \textbf{Section \ref{sec:image_reconstruction_error}}), performance of abnormality segmentation (see \textbf{Section \ref{sec:segmentation_performance}}), leakiness of feature decomposition (see \textbf{Section \ref{sec:leakiness_of_feature_decomposition}}), and compactness of the codebooks (see \textbf{Section \ref{sec:compactness_of_codebooks}}). As shown in {\bf Fig. \ref{fig:image_recon_results}} and {\bf Fig. \ref{fig:segmentation_results}}, the performances of both image reconstruction and segmentation were improved as the resolution of the latent representation increased, which is demonstrated by the reduced reconstruction errors and increased Dice coefficients, respectively. Since the distribution matching is expected to impose some regularization effects on the codebooks, the performance of image reconstruction and segmentation could be degraded by training with distribution matching; however, the data also indicate that it did not cause any negative effect on both image reconstruction and segmentation performance. For the leakiness evaluated in {\bf Fig. \ref{fig:leakiness_results}}, without the distribution matching, the tendency that the normal-appearing reconstructions unintentionally contained distinguishable features of abnormality became apparent when the latent resolution increased. For example, this can be evident at $\mathrm{FDN}_{32 \times 32}$, where PPV for the origin of diseased images of the classification network trained based on normal-appearing reconstructions (see the orange bar in {\bf Fig. \ref{fig:leakiness_results}}) was as high as those of the classification network trained based on entire reconstructions (see the yellow and gray bars in {\bf Fig. \ref{fig:leakiness_results}}). However, the leakiness of abnormal features could be mitigated to some extent by introducing distribution matching. Note that PPVs for the origin of diseased images of the classification network trained on normal-appearing reconstructions (see the orange bars in {\bf Fig. \ref{fig:leakiness_results}}) consistently decreased by imposing distribution matching (see the blue bars in {\bf Fig. \ref{fig:leakiness_results}}). Regarding the compactness of the codebooks shown in {\bf Table \ref{tab:compactness_of_codebooks}}, models with lower latent resolutions showed higher value. Besides, it is noteworthy that distribution matching had the effect of rendering the codebooks more compact with the higher ratio of insignificant code vectors, which can be observed except for the abnormal anatomy codes of $\mathrm{FDN}_{16 \times 16}$ with distribution matching. See {\bf \ref{app:distribution_of_norm}} for the distribution of norms of code vectors in each configuration, which also supports that models with low latent space resolution had more code vectors with small norms.

In summary, there is a tradeoff between the performance of image reconstruction and segmentation and leakiness of feature decomposition and compactness of the codebooks according to the spatial resolution of latent representations. Note that, when the length and width of the latent space doubles (e.g., from $8 \times 8$ to $16 \times 16$), the computational complexity of the code vectors quadruples. Hereinafter, we selected $\mathrm{FDN}_{8 \times 8}$ with distribution matching for the following evaluation because the model exhibited intermediate characteristics, enabling good feature decomposition and simple latent representation with compact codebooks, which will be favorable at the time of similarity calculation. 

\subsection{Assessment of the binarization process}
\label{sec:assessment_of_binarization}

\begin{figure}[t]
  \centering
  \includegraphics[]{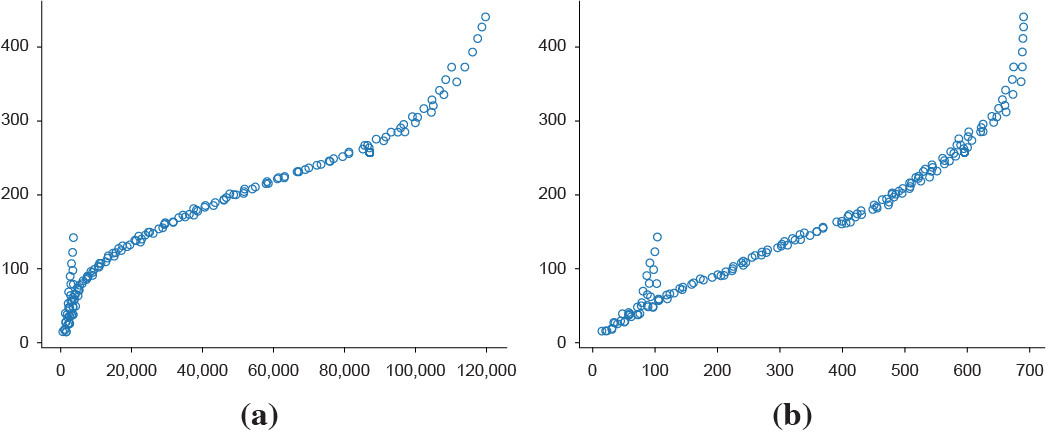}
  \caption{\textbf{Relationship between Hamming distance and Euclidean distance.} The distances between one code vector and all other code vectors were calculated by Hamming distance (horizontal axis) and Euclidean distance (vertical axis). \textbf{(a)} Distance relationship between the continuous codebook and binarized codebook before optimization. \textbf{(b)} The same relationship after the optimization of the binarized codebook length. Note that not only the nearest relationship but also the global relationship is maintained through the optimization process.}
  \label{fig:binarization}
\end{figure}

Using $\mathrm{FDN}_{8 \times 8}$ with distribution matching as the feature decomposing network, we validated the process of the binarization of code vectors (see \textbf{Section \ref{sec:validity_of_binarization_process}}). Based on the techniques described in \textbf{Section \ref{sec:optimization_of_binarized_vector_length}}, the length of the optimized binarized vector was shortened from $13,816$ to 789 for the normal anatomy codes and 292 for the abnormal anatomy codes. The compression ratios $\frac{E^*}{E}$ for the normal and abnormal anatomy codes were 0.60\% and 0.22\%, respectively. {\bf Fig. \ref{fig:binarization}} demonstrates an example result on the relationship between Hamming distance $D_\mathrm{H}$ and Euclidean distance $D_\mathrm{E}$ before ({\bf Fig. \ref{fig:binarization}a}) and after ({\bf Fig. \ref{fig:binarization}b}) optimization of the vector length. Note that, even though the optimization process considered only the nearest neighbor relationship, the global distance relationship was also maintained. The concordances of the top 1, 5, and 10 closest relationships between the Hamming distance calculation using the optimized binarized codebook $\bm{b}^*$ and the Euclidean distance calculation using the continuous codebook $\bm{e}$ were 0.81, 0.89, and 0.90 for normal anatomy codes and 0.81, 0.88, and 0.90 for abnormal anatomy codes, respectively. Notably, the relationships between the binarized codebook $\bm{b}$ before the optimization and the continuous codebook $\bm{e}$ were at similar levels, demonstrating that the concordances of top 1, 5, and 10 closest relationships were 0.82, 0.86, and 0.89 for normal anatomy codes and 0.84, 0.87, and 0.89 for abnormal anatomy codes, respectively. Therefore, the optimization process of the binarized vector length did not hinder the distance relationship with respect to the original Euclidean distance. Hamming distance calculation using the optimized binarized codebook $\bm{b}^*$ reduced the computational time by 48.3\% and 64.5\% for normal and abnormal anatomy codes, respectively, compared to the Euclidean distance calculation using the continuous codebook $\bm{e}$ (see \textbf{\ref{app:comparison_of_computational_time}}).

\subsection{Retrieval results based the decomposed latent codes}
\label{sec:result_query_by_image}

\begin{figure*}[t]
  \centering
  \includegraphics[]{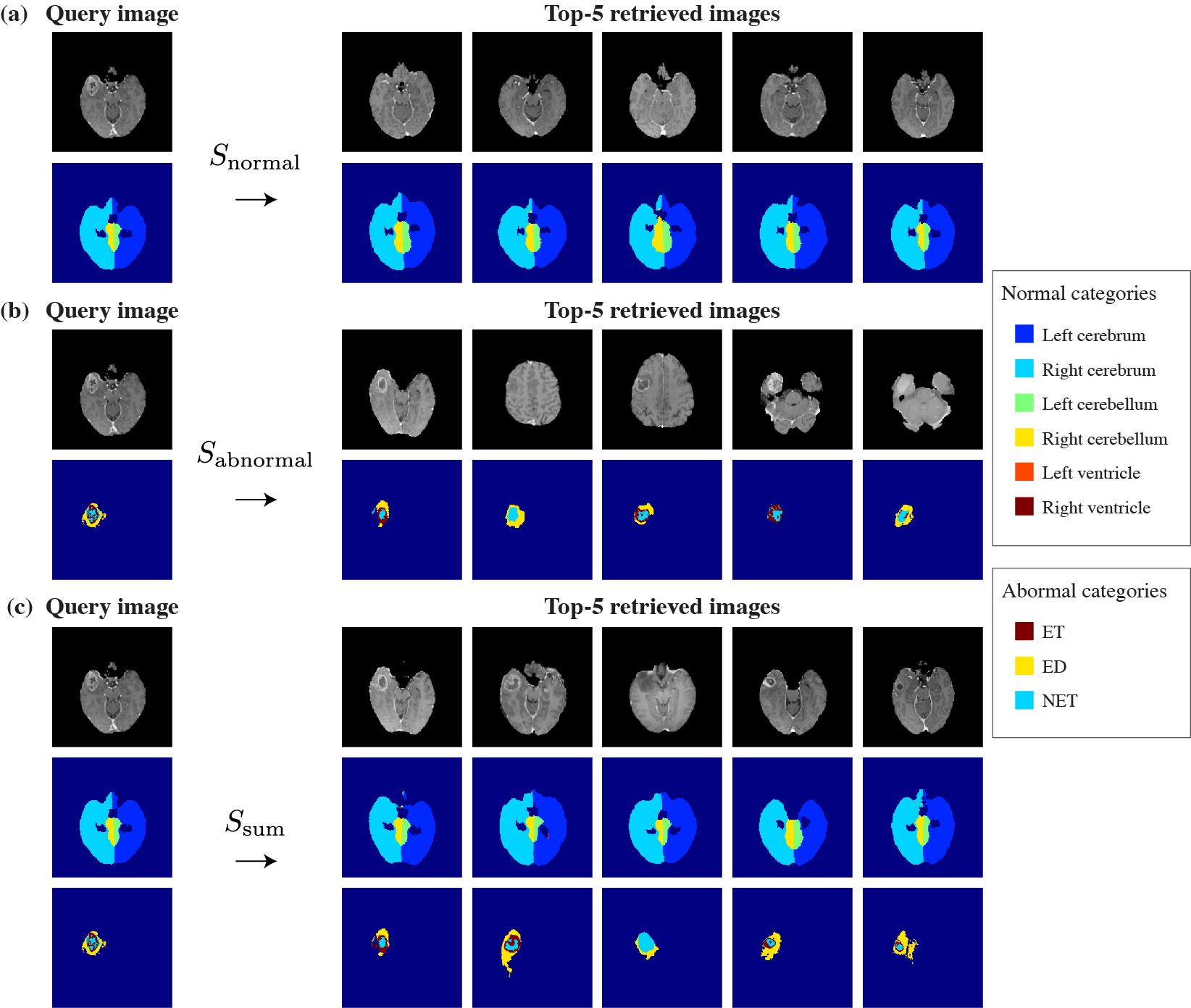}
  \caption{\textbf{Example results of CBIR based on the decomposed latent codes.} Image retrieval was performed based on volume, comparing the closest images from each MR volume. Each similarity was calculated according to Euclidean distance $D_\mathrm{E}$ based on the continuous codebooks $\bm{e}$. Retrieved images are arranged from left to right, starting with the closest one to the query vector. ({\bf a}) Similarity calculation based on normal anatomy codes $S_\mathrm{normal}$ retrieved images with similar normal anatomical labels, irrespective of gross abnormalities. Query and retrieved images with ground-truth labels of the normal anatomical categories are shown. ({\bf b}) Similarity calculation based on abnormal anatomy codes $S_\mathrm{abnormal}$ retrieved images with similar abnormal anatomical labels. Query and retrieved images with ground-truth labels of the abnormal anatomical categories are shown. Note the variety of normal anatomical contexts of the retrieved images. ({\bf c}) Similarity calculation using $S_\mathrm{sum}$ retrieved images with combined features, as shown by the similar patterns of both normal and abnormal anatomical labels among the retrieved images. Query and retrieved images with ground-truth labels of the normal and abnormal anatomical categories are shown. ET, Gd-enhancing tumor; ED, peritumoral edema; NET, necrotic and non-enhancing tumor core.}
  \label{fig:query_by_image}
\end{figure*}

\begin{table}[t]
\centering
\resizebox{0.8\linewidth}{!}{%
\begin{tabular}{@{}lccc@{}}
\toprule
\textbf{Distance measurements} & \boldsymbol{$S_\mathrm{normal}$} & \boldsymbol{$S_\mathrm{abnormal}$} & \boldsymbol{$S_\mathrm{sum}$}\\ \midrule
$D_\mathrm{E}$                 & $0.50 \pm 0.16$              & $0.28 \pm 0.12$                        & $0.20 \pm 0.16$\\
$D_\mathrm{A}$                 & $0.47 \pm 0.19$              & $0.27 \pm 0.11$                        & $0.30 \pm 0.11$\\
$D_\mathrm{H}$                 & $0.48 \pm 0.17$              & $0.28 \pm 0.12$                        & $0.23 \pm 0.15$\\
Brutal search                  & $0.62 \pm 0.16$              & $0.37 \pm 0.08$                        & $0.42 \pm 0.08$\\ \bottomrule
\end{tabular}
}
\caption{\textbf{Quantitative evaluation of the image retrieval.} Mean $\pm$ standard deviation of the averaged Dice coefficients for ground-truth labels with respect to the semantic components exploited by the similarity evaluation is shown.} 
\label{tab:quantitative_evaluation}
\end{table}

An example of the CBIR results showing the top 5 images with the closest latent codes based on $S_\mathrm{normal}$, $S_\mathrm{abnormal}$, and $S_\mathrm{sum}$ is presented in {\bf Fig. \ref{fig:query_by_image}}. Similarity calculation based on normal anatomy codes $S_\mathrm{normal}$ retrieved images with similar normal anatomical labels irrespective of gross abnormalities ({\bf Fig. \ref{fig:query_by_image}a}). Similarity calculation based on abnormal anatomy codes $S_\mathrm{abnormal}$ retrieved images with similar abnormal anatomical labels ({\bf Fig. \ref{fig:query_by_image}b}). Note that the variety of normal anatomical contexts of the retrieved images accompanied similar abnormal lesions. In the similarity retrieval using $S_\mathrm{sum}$, as formulated in Eq. (\ref{eq:dsum}), the similarity measurements between the normal and abnormal anatomy codes were summed. As shown in {\bf Fig. \ref{fig:query_by_image}c}, the whole imaging features of the retrieved images using $S_\mathrm{sum}$ resemble those of the query image. In this example, Euclidean distance $D_\mathrm{E}$ was employed based on the continuous codebooks $\bm{e}$. 

According to the method described in \textbf{Section \ref{sec:eval_query_by_image}}, we performed top 10 nearest neighbor search between query images and reference dataset for quantitative evaluation. Here, the query images are the slices with the largest tumor area in each MR volume, and the reference dataset contains the rest of images in the test dataset, except for the images in the same MR volume of each query image. The similarity measurements utilized $S_\mathrm{normal}$, $S_\mathrm{abnormal}$, and $S_\mathrm{sum}$. Mean Dice coefficients were assessed according to the semantics exploited in the similarity measurement: Dice coefficients between ground-truth labels of the normal anatomical categories were averaged when using $S_\mathrm{normal}$, those between ground-truth labels of the abnormal anatomical categories were averaged when using $S_\mathrm{abnormal}$, and those between ground-truth labels of the normal anatomical categories and those between ground-truth labels of the abnormal anatomical categories were averaged when using $S_\mathrm{sum}$. The three types of distance calculations, Euclidean distance $D_\mathrm{E}$, angular distance $D_\mathrm{A}$, and Hamming distance $D_\mathrm{H}$, were calculated for each similarity measurement. The results are summarized in {\bf Table \ref{tab:quantitative_evaluation}}. The technical upper bound was provided by the brutal search, which retrieved images to maximize Dice coefficients directly computing the overlap between ground-truth labels. Note that the brutal search is performed based on the ground-truth labels for all query and reference images, which is usually not given in real clinical practice. 

\section{Discussion and conclusions}
\label{sec:discussion}

Comparative diagnostic reading is critical in correct diagnosis by comparing an image of the condition to be diagnosed with corresponding normal images without abnormal findings or images that contain similar abnormal findings. To the best of our knowledge, this is the first study that proposes a deep-learning-based algorithm specifically designed to support the comparative diagnostic reading. The fundamental contribution of our study is to extend the idea of disentangled representation into a CBIR application in medical imaging. By leveraging the feature decomposing network, a medical image can be decomposed into a normal and abnormal anatomy code, each of which represents the targeted semantic component in the image. Note that the codebooks located at the bottom of the network allows decomposed latent codes to be manipulable for the CBIR framework, which can switch the semantic components to be focused in the retrieval according to users' purposes. 

The proposed CBIR framework demonstrated notable results in both qualitative and quantitative evaluation (see \textbf{Section \ref{sec:result_query_by_image}}). However, because of the two-staged approach to establish the CBIR framework, there are possible error origins for the final image retrieval performance, such as errors in image reconstruction, segmentation, and vector quantization in the codebooks. Moreover, the leakiness of abnormal imaging features into the normal anatomical codes can hinder retrieval performance such that the retrieval based on $D^\mathrm{normal}$ unintentionally accompanies images with significant amount of abnormalities, which can be alleviated by distribution matching. Because it is quite overloaded as an experiment, directly comparing the image retrieval performance according to the different configurations of the feature decomposing network could not be performed. Instead, we precisely evaluated these possible origins of errors (see \textbf{Section \ref{sec:comparison_between_several_configuration}}) and selected $\mathrm{FDN}_{8 \times 8}$ trained with distribution matching for reporting the CBIR performance owing to its preferable features for the image retrieval.

Even though there is no benchmark available for the image retrieval based on the dataset, we consider that the performance difference from the brutal search (see \textbf{Table \ref{tab:quantitative_evaluation}}) can be acceptable for clinical use. Under the condition that the leakiness of abnormal imaging features is controlled, it is quite natural to speculate that the performance of the image reconstruction and segmentation has a positive relationship between the image retrieval performance using the decomposed latent codes. Therefore, if higher retrieval performance is desired and computational resources are sufficient, a model with higher latent space resolution can be used. Because the brutal search is an ideal setting, where ground-truth labels are given to all query and reference images, our CBIR framework is more versatile and useful, and furthermore, different configurations can be used for different situations. 

Distance calculations using Euclidean distance $D_\mathrm{E}$, angular distance $D_\mathrm{A}$, and Hamming distance $D_\mathrm{H}$ can be implemented according to different situations, taking into account the tradeoff between accuracy and computational efficiency at the time of similarity search. Note that, with respect to $S_\mathrm{sum}$, angular distance calculation $D_\mathrm{A}$ provided a higher value than others. This can be because the magnitude of the distance values is not normalized in the Euclidean distance calculation $D_\mathrm{E}$ and Hamming distance calculation $D_\mathrm{H}$, making it difficult to evenly weight the two terms, $S_\mathrm{normal}(q, r)$ and $S_\mathrm{abnormal}(q, r)$, in Eq. (\ref{eq:dsum}). It is also noteworthy that, even though the proposed binarization technique is simple, Hamming distance calculation $D_\mathrm{H}$ approximated the Euclidean distance calculation $D_\mathrm{E}$ with notable accuracy (see \textbf{Section \ref{sec:assessment_of_binarization}}). Since the picture archiving and communication systems in hospitals usually contain a huge amount of medical images, our proposal of the binary hashing based on the discrete latent codes can also be effective even in the context of large-scale image search. 

One may argue that the distribution matching should be applied in reconstructed images instead of latent codes. It seems to be one alternative method to decompose features of medical images, but there are some concerns. One is an issue called ``posterior collapse,'' which is caused by a powerful decoder ignoring latent codes, which can be observed in many VAE models \citep{oord2017neural}. Because our primary purpose is to acquire good latent representation that can be faithfully representative for corresponding imaging features, we intentionally avoided imposing additional learning objective for decoders. Another concern is that the distribution matching for images with a resolution of $256 \times 256$ was computationally expensive and required a long time for training of feature decomposing networks. We also encountered a more unstable training process, which is one of the intrinsic problems of GANs. As shown in \textbf{Fig. \ref{fig:comparison_reconstructions}}, where the constraints on the latent space are reflected in the differences in imaging features in the reconstructed images, we consider that distribution matching on the latent spaces should be more appropriate when it comes to CBIR utilizing latent codes. 

We found that SPADE modules can effectively propagate the semantic layout obtained at the final layer of the segmentation decoder into the image reconstruction process of the image decoder. The success of the conditional image generation of the image decoder depending on the collateral input through the SPADE modules can also be observed in \textbf{Fig. \ref{fig:leakiness_results}}, where PPVs for the origin from diseased images of the classification network trained using the entire reconstructions were consistently $>$ 0.9 (see yellow and gray bars in \textbf{Fig. \ref{fig:leakiness_results}}). However, future technological challenges may lie in this regard. Ideally, the latent codes representing normal and abnormal semantic components of medical images should be distributed in a decomposable manner in a single space where they can be linearly computed with each other. Because the current study employed an architecture that holds two separated latent spaces for decomposed latent codes, the similarity calculation according to the whole imaging feature (Eq. (\ref{eq:dsum})) might be deemed arbitrary. Even so, because simple autoencoders can be sufficient to calculate the similarity based on whole imaging features, we believe that our proposal is innovative to enable the CBIR to selectively utilize either normal or abnormal components of medical images to support comparative diagnostic reading.

\appendix

\section{Optimization algorithm for the binarized vector length}
\label{app:optimization_algorithm}
\setcounter{algocf}{0}
\renewcommand{\thealgocf}{A.\arabic{algocf}}

Optimization algorithm for the binarized vector length is shown in {\bf Algorithm \ref{alg:optimizing_algorithm}}.

\begin{algorithm}[t]
\SetAlgoLined
  \DontPrintSemicolon
  $\bm{b} = \{b_k | k = 1, \dots, K\} \in \mathbb{R}^{K \times E}$: initial codebook\\
  $\bm{b}^* = \{b_k^* | k = 1, \dots, K\} \in \mathbb{R}^{K \times E^*}$: optimized codebook\\
  $E =$ $K \choose 2$: the initial length of code vectors\\
  $D_\mathrm{H}$: Hamming distance based on code vectors\\
  $D^\prime_\mathrm{H}$: Hamming distance based on code vectors with an element removed\\
  \SetKwFunction{FMain}{Optimize}
  \SetKwProg{Pn}{Function}{:}{\KwRet $\bm{b}^*$}
  \Pn{\FMain{$\bm{b}$, $E$}}{
  \For{e = 1, $\dots$, E}{
       \For{i = 1, $\dots$, K}{
       $b_i \leftarrow \bm{b}[i]$\\
       $b_i^\prime \leftarrow$ remove $e$-th element from $b_i$\\
       \For{j = i + 1, $\dots$, K}{
           $b_j \leftarrow \bm{b}[j]$\\
           $b_j^\prime \leftarrow$ remove $e$-th element from $b_j$\\
           Compute $D_\mathrm{H} (b_i, b_j)$ to obtain a set of the pairs of (i, j) to minimize the distance.\\
           Compute $D^\prime_\mathrm{H} (b_i^\prime, b_j^\prime)$ to obtain a set of the pairs of (i, j) to minimize the distance.\\
       }
       \uIf{$\argmin_{\{(i, j)\}} D_\mathrm{H} \neq \argmin_{\{(i, j)\}} D^\prime_\mathrm{H}$}{
            \it{is\_deletable} $\leftarrow$ {\it False}\\
            break\\
        }
       \Else{
            continue\\
       }
       }
       }
       
      \uIf{\it{is\_deletable} $=$ {\it True}}{
        Remove $e$-th element from $\bm{b}$.\\
        $\bm{b}^\prime \leftarrow \bm{b}$\\
        $E^\prime \leftarrow E - 1$\\
        \FMain{$\bm{b}^\prime$, $E^\prime$}\\
      }
      \Else{
        continue\\
      }

  }
\caption{Optimization of the binarized vector length}
\label{alg:optimizing_algorithm}
\end{algorithm}

\section{Annotation process for the MICCAI BraTS validation and test dataset}
\label{app:annotation_process}
\setcounter{figure}{0}   

\subsection{Segmentation for the normal anatomical categories}

For the three datasets obtained from the 2019 BraTS Challenge \citep{6975210brats, Bakas2017, TCGAGBM, TCGALGG}, MICCAI\_BraTS\_Training, MICCAI\_BraTS\_Validation, and MICCAI\_BraTS\_Testing, we assigned six normal anatomical labels (left cerebrum, right cerebrum, left cerebellum, right cerebellum, left ventricle, and right ventricle) using the software BrainSuite 19a \citep{SHATTUCK2002129}. Briefly, the software provides an image processing pipeline, including skull stripping, correction for image nonuniformities, and voxel-level classification of neuroanatomical structures. Among the automatically extracted structures, we obtained six normal anatomical categories for the present study. 

\subsection{Segmentation for the abnormal anatomical categories}

Among the three datasets from the 2019 BraTS Challenge, only MICCAI\_BraTS\_Training had three types of abnormality labels (ET, ED, and TC). To segment the tumor regions in other datasets, MICCAI\_BraTS\_Validation and MICCAI\_BraTS\_Testing, we first trained a 3D-based segmentation network whose basic implementation is almost similar to that of a previous study \citep{myronenko2020robust}. The training and validation of the segmentation network were conducted using fivefold cross-validation in MICCAI\_BraTS\_Training, demonstrating that the average running Dice coefficients in the validation splits at the epoch 300 for ET, ED, and TC were 0.68, 0.74, and 0.79, respectively. Then, outputs from the five models that were separately trained in each split were averaged for the prediction of the abnormal labels in MICCAI\_BraTS\_Validation and MICCAI\_BraTS\_Testing. Our source code and training settings for this process are available on public (https://github.com/Kaz-K/robust-glioma-segmentation). Lastly, expert radiologists visually reviewed and manually edited the segmentation results in MICCAI\_BraTS\_Validation and MICCAI\_BraTS\_Testing.

\section{Detailed architecture of the feature decomposing network}
\label{app:detailed_architecture}
\setcounter{figure}{0}   

Detailed architecture of the feature decomposing network with the latent size of $8 \times 8$, $\mathrm{FDN}_{8 \times 8}$, is presented in {\bf Table \ref{tab:encoder_architecture}} and {\bf Table \ref{tab:decoder_architecture}}. There are repeated structures with the same kernel size, such as (ResBlock + StridedConv) for the encoder and [Upsample + (SPADE-)ResBlock] for the decoders, the number of which is adjusted according to the size of the latent space defined in each configuration. The architecture of the critic network working for the distribution matching is also shown in {\bf Table \ref{tab:critic_architecture}}. In all networks, LeakyReLU \citep{xu2015empirical} is used for the activation function. 

\begin{table}[H]
\renewcommand\thetable{C.1} 
\centering
\caption{\textbf{Detailed architecture of the encoder.}}
\resizebox{\linewidth}{!}{%
\begin{tabular}{lcc}
\hline
\textbf{Module}                                                     & \textbf{Activation}                                                                                                                                                                                                         & \textbf{Output shape}                                                                                                               \\ \hline
\begin{tabular}[c]{@{}l@{}}Input image\\ Conv\\ StridedConv\end{tabular} & \begin{tabular}[c]{@{}c@{}}\\$\begin{bmatrix} 3 \times 3 & 32 \end{bmatrix}$\\ $\begin{bmatrix} 3 \times 3 & 64 \end{bmatrix}$\end{tabular}                                                                                                                        & \begin{tabular}[c]{@{}c@{}}$4 \times 256 \times 256$\\ $32 \times 256 \times 256$\\ $64 \times 128 \times 128$\end{tabular} \\ \hline
\begin{tabular}[c]{@{}l@{}}ResBlock\\ \\ StridedConv \\ Split\end{tabular}       & \begin{tabular}[c]{@{}c@{}}$\begin{bmatrix} 3 \times 3 & 64 \\ 3 \times 3 & 64 \end{bmatrix}$\\ $\begin{bmatrix} 3\times 3 & 128\end{bmatrix}$\\ -\end{tabular}                                                                                                     & \begin{tabular}[c]{@{}c@{}}$64 \times 128 \times 128$\\ \\ $128 \times 64 \times 64$ \\ $128 \times 64 \times 64, 128 \times 64 \times 64$\end{tabular}                            \\ \hline
\begin{tabular}[c]{@{}l@{}}ResBlock\\ \\ StridedConv\end{tabular}       & \begin{tabular}[c]{@{}c@{}}$\begin{bmatrix} 3 \times 3 & 128 \\ 3 \times 3 & 128 \end{bmatrix}, \begin{bmatrix} 3 \times 3 & 128 \\ 3 \times 3 & 128 \end{bmatrix}$\\ $\begin{bmatrix} 3\times 3 & 128\end{bmatrix}, \begin{bmatrix} 3\times 3 & 128\end{bmatrix}$\end{tabular}                                                                                                   & \begin{tabular}[c]{@{}c@{}}$128 \times 64 \times 64, 128 \times 64 \times 64$\\ \\ $128 \times 32 \times 32, 128 \times 32 \times 32$\end{tabular}                            \\ \hline
\begin{tabular}[c]{@{}l@{}}ResBlock\\ \\ StridedConv\end{tabular}       & \begin{tabular}[c]{@{}c@{}}$\begin{bmatrix} 3 \times 3 & 128 \\ 3 \times 3 & 128 \end{bmatrix}, \begin{bmatrix} 3 \times 3 & 128 \\ 3 \times 3 & 128 \end{bmatrix}$\\ $\begin{bmatrix} 3\times 3 & 128\end{bmatrix}, \begin{bmatrix} 3\times 3 & 128\end{bmatrix}$\end{tabular}                                                                                                   & \begin{tabular}[c]{@{}c@{}}$128 \times 32 \times 32, 128 \times 32 \times 32$\\ \\ $128 \times 16 \times 16, 128 \times 16 \times 16$\end{tabular}                            \\ \hline
\begin{tabular}[c]{@{}l@{}}ResBlock\\ \\ StridedConv\end{tabular}       & \begin{tabular}[c]{@{}c@{}}$\begin{bmatrix} 3 \times 3 & 128 \\ 3 \times 3 & 128 \end{bmatrix}, \begin{bmatrix} 3 \times 3 & 128 \\ 3 \times 3 & 128 \end{bmatrix}$\\ $\begin{bmatrix} 3\times 3 & 128\end{bmatrix}, \begin{bmatrix} 3\times 3 & 128\end{bmatrix}$\end{tabular}                                                                                                   & \begin{tabular}[c]{@{}c@{}}$128 \times 16 \times 16, 128 \times 16 \times 16$\\ \\ $128 \times 8 \times 8, 128 \times 8 \times 8$\end{tabular}                            \\ \hline
\begin{tabular}[c]{@{}l@{}}ResBlock\\ \\ ResBlock\end{tabular}       & \begin{tabular}[c]{@{}c@{}}$\begin{bmatrix} 3 \times 3 & 128 \\ 3 \times 3 & 128 \end{bmatrix}, \begin{bmatrix} 3 \times 3 & 128 \\ 3 \times 3 & 128 \end{bmatrix}$\\ $\begin{bmatrix} 3 \times 3 & 64 \\ 3 \times 3 & 64 \end{bmatrix}, \begin{bmatrix} 3 \times 3 & 64 \\ 3 \times 3 & 64 \end{bmatrix}$\end{tabular}                                                                                                   & \begin{tabular}[c]{@{}c@{}}$128 \times 8 \times 8, 128 \times 8 \times 8$\\ \\ $64 \times 8 \times 8, 64 \times 8 \times 8$\end{tabular}                            \\ \hline
  \end{tabular}
}
\label{tab:encoder_architecture}
\end{table} 

\begin{table}[H]
\renewcommand\thetable{C.2} 
\centering
\caption{\textbf{Detailed architecture shared between the image and segmentation decoders.}}
\resizebox{0.8\linewidth}{!}{%
\begin{tabular}{lcc}
\hline
\textbf{Module}                                                & \textbf{Activation}                                                                                                                                                     & \textbf{Output shape}                                                                                       \\ \hline
Latent representation                                           & -                                                                                                                                                              & $64 \times 8 \times 8$                                                                            \\ \hline
ResBlock                                                      & $\begin{bmatrix} 3 \times 3 & 128 \\ 3 \times 3 & 128 \end{bmatrix}$                                                                                           & $128 \times 8 \times 8$                                                                            \\ \hline
\begin{tabular}[c]{@{}l@{}}Upsample\\ (SPADE-)ResBlock \\ \\ \end{tabular} & \begin{tabular}[c]{@{}c@{}}$\begin{bmatrix} 3 \times 3 & 128\end{bmatrix}$\\ $\begin{bmatrix} 3 \times 3 & 128 \\ 3 \times 3 & 128 \end{bmatrix}$\end{tabular} & \begin{tabular}[c]{@{}c@{}}$128 \times 16 \times 16$\\ $128 \times 16 \times 16$ \\ \\ \end{tabular}   \\ \hline
\begin{tabular}[c]{@{}l@{}}Upsample\\ (SPADE-)ResBlock \\ \\ \end{tabular}  & \begin{tabular}[c]{@{}c@{}}$\begin{bmatrix} 3 \times 3 & 128\end{bmatrix}$\\ $\begin{bmatrix} 3 \times 3 & 128 \\ 3 \times 3 & 128 \end{bmatrix}$\end{tabular} & \begin{tabular}[c]{@{}c@{}}$128 \times 32 \times 32$\\ $128 \times 32 \times 32$ \\ \\ \end{tabular}   \\ \hline
\begin{tabular}[c]{@{}l@{}}Upsample\\ (SPADE-)ResBlock \\ \\ \end{tabular}  & \begin{tabular}[c]{@{}c@{}}$\begin{bmatrix} 3 \times 3 & 128\end{bmatrix}$\\ $\begin{bmatrix} 3 \times 3 & 128 \\ 3 \times 3 & 128 \end{bmatrix}$\end{tabular}    & \begin{tabular}[c]{@{}c@{}}$128 \times 64 \times 64$\\ $128 \times 64 \times 64$ \\ \\ \end{tabular}     \\ \hline
\begin{tabular}[c]{@{}l@{}}Upsample\\ (SPADE-)ResBlock \\ \\ \end{tabular}  & \begin{tabular}[c]{@{}c@{}}$\begin{bmatrix} 3 \times 3 & 64\end{bmatrix}$\\ $\begin{bmatrix} 3 \times 3 & 64 \\ 3 \times 3 & 64 \end{bmatrix}$\end{tabular}    & \begin{tabular}[c]{@{}c@{}}$64 \times 128 \times 128$\\ $64 \times 128 \times 128$ \\ \\ \end{tabular} \\ \hline
\begin{tabular}[c]{@{}l@{}}Upsample\\ (SPADE-)ResBlock \\ \\ \end{tabular}  & \begin{tabular}[c]{@{}c@{}}$\begin{bmatrix} 3 \times 3 & 32\end{bmatrix}$\\ $\begin{bmatrix} 3 \times 3 & 32 \\ 3 \times 3 & 32 \end{bmatrix}$\end{tabular}    & \begin{tabular}[c]{@{}c@{}}$32 \times 256 \times 256$\\ $32 \times 256 \times 256$ \\ \\ \end{tabular} \\ \hline
Conv                                                            & $\begin{bmatrix} 1 \times 1 & 1\end{bmatrix}$                                                                                                                  & $1 \times 256 \times 256$                                                                          \\ \hline
\end{tabular}%
}
\label{tab:decoder_architecture}
\end{table} 

\begin{table}[H]
\renewcommand\thetable{C.3} 
\centering
\caption{\textbf{Detailed architecture of the critic network.}}
\resizebox{0.8\linewidth}{!}{%
\begin{tabular}{lcc}
\hline
\textbf{Module}                                                & \textbf{Activation}                                                                                                                                                     & \textbf{Output shape}                                                                                       \\ \hline
Latent representation                                           & -                                                                                                                                                             & $64 \times 8 \times 8$                                                                            \\ \hline
StridedConv                                                     & $\begin{bmatrix} 4 \times 4 & 64 \end{bmatrix}$                                                                                                               & $64 \times 4 \times 4$                                                                            \\ \hline
StridedConv                                                     & $\begin{bmatrix} 4 \times 4 & 64 \end{bmatrix}$                                                                                                               & $64 \times 2 \times 2$                                                                            \\ \hline
StridedConv                                                     & $\begin{bmatrix} 4 \times 4 & 64 \end{bmatrix}$                                                                                                               & $64 \times 1 \times 1$                                                                            \\ \hline
Conv                                                            & $\begin{bmatrix} 1 \times 1 & 64\end{bmatrix}$                                                                                                                & $64 \times 1 \times 1$                                                                          \\ \hline
Conv                                                            & $\begin{bmatrix} 1 \times 1 & 1\end{bmatrix}$                                                                                                                 & $1 \times 1 \times 1$                                                                          \\ \hline
\end{tabular}%
}
\label{tab:critic_architecture}
\end{table} 

\section{Training details of the classification network}
\label{app:detailed_architecture_classifier}
\setcounter{figure}{0}

The architecture of the classification network is shown in {\bf Fig. \ref{tab:classifier_network}}. For the activation function, ReLU \citep{10.5555/3104322.3104425} is used. The input for the classification network was normal-appearing images $\hat{\bm{x}}^-$, and its output was binary vectors corresponding to the presence or absence of abnormality in the original images $\bm{x}$. Binary cross-entropy loss was evaluated for the prediction error to train the network. Other hyperparameters were shared across the configurations as follows: batch size = 112, number of training epochs = 100, learning late = $1.0 \times 10^{-4}$, and weight decay = $1.0 \times 10^{-5}$.

\begin{table}[H]
\renewcommand\thetable{D.1} 
\centering
\caption{\textbf{Detailed architecture of the classification network}}
\resizebox{0.7\linewidth}{!}{%
\begin{tabular}{lcc}
\hline
\textbf{Module}                                                & Activation                                                                                                                                                     & Output shape                                                                                       \\ \hline

\begin{tabular}[c]{@{}l@{}}Input image \\ Conv \\ Conv \\ MaxPool \\ \end{tabular}  & \begin{tabular}[c]{@{}c@{}}\\ $\begin{bmatrix} 3 \times 3 & 32\end{bmatrix}$\\ $\begin{bmatrix} 3 \times 3 & 32\end{bmatrix}$ \\ - \end{tabular} & \begin{tabular}[c]{@{}c@{}}$4 \times 256 \times 256$ \\ $32 \times 256 \times 256$\\ $32 \times 256 \times 256$ \\ $32 \times 128 \times 128$ \\ \end{tabular}   \\ \hline

\begin{tabular}[c]{@{}l@{}}Conv \\ Conv \\ MaxPool \\ \end{tabular}  & \begin{tabular}[c]{@{}c@{}}$\begin{bmatrix} 3 \times 3 & 64\end{bmatrix}$\\ $\begin{bmatrix} 3 \times 3 & 64\end{bmatrix}$ \\ - \end{tabular} & \begin{tabular}[c]{@{}c@{}}$64 \times 128 \times 128$\\ $64 \times 128 \times 128$ \\ $64 \times 64 \times 64$ \\ \end{tabular}   \\ \hline

\begin{tabular}[c]{@{}l@{}}Conv \\ Conv \\ Conv \\ MaxPool \\ \end{tabular}  & \begin{tabular}[c]{@{}c@{}}$\begin{bmatrix} 3 \times 3 & 128 \end{bmatrix}$\\ $\begin{bmatrix} 3 \times 3 & 128 \end{bmatrix}$\\ $\begin{bmatrix} 3 \times 3 & 128 \end{bmatrix}$ \\ - \end{tabular} & \begin{tabular}[c]{@{}c@{}}$128 \times 64 \times 64$\\ $128 \times 64 \times 64$\\ $128 \times 64 \times 64$ \\ $128 \times 32 \times 32$ \\ \end{tabular}   \\ \hline

\begin{tabular}[c]{@{}l@{}}Conv \\ Conv \\ Conv \\ MaxPool \\ \end{tabular}  & \begin{tabular}[c]{@{}c@{}}$\begin{bmatrix} 3 \times 3 & 256 \end{bmatrix}$\\ $\begin{bmatrix} 3 \times 3 & 256 \end{bmatrix}$\\ $\begin{bmatrix} 3 \times 3 & 256 \end{bmatrix}$ \\ - \end{tabular} & \begin{tabular}[c]{@{}c@{}}$256 \times 32 \times 32$\\ $256 \times 32 \times 32$\\ $256 \times 32 \times 32$ \\ $256 \times 16 \times 16$ \\ \end{tabular}   \\ \hline

\begin{tabular}[c]{@{}l@{}}Conv \\ Conv \\ Conv \\ MaxPool \\ \end{tabular}  & \begin{tabular}[c]{@{}c@{}}$\begin{bmatrix} 3 \times 3 & 512 \end{bmatrix}$\\ $\begin{bmatrix} 3 \times 3 & 512 \end{bmatrix}$\\ $\begin{bmatrix} 3 \times 3 & 512 \end{bmatrix}$ \\ - \end{tabular} & \begin{tabular}[c]{@{}c@{}}$512 \times 16 \times 16$\\ $512 \times 16 \times 16$\\ $512 \times 16 \times 16$ \\ $512 \times 8 \times 8$ \\ \end{tabular}   \\ \hline

\begin{tabular}[c]{@{}l@{}}Conv \\ Conv \\ Conv \\ MaxPool \\ \end{tabular}  & \begin{tabular}[c]{@{}c@{}}$\begin{bmatrix} 3 \times 3 & 512 \end{bmatrix}$\\ $\begin{bmatrix} 3 \times 3 & 512 \end{bmatrix}$\\ $\begin{bmatrix} 3 \times 3 & 512 \end{bmatrix}$ \\ - \end{tabular} & \begin{tabular}[c]{@{}c@{}}$512 \times 8 \times 8$\\ $512 \times 8 \times 8$\\ $512 \times 8 \times 8$ \\ $512 \times 4 \times 4$ \\ \end{tabular}   \\ \hline

\begin{tabular}[c]{@{}l@{}}Linear \\ Linear \\ \end{tabular}  & \begin{tabular}[c]{@{}c@{}}$\begin{bmatrix} 8192 & 256 \end{bmatrix}$\\ $\begin{bmatrix} 256 & 1 \end{bmatrix}$ \end{tabular} & \begin{tabular}[c]{@{}c@{}} $256$ \\ $1$ \end{tabular}   \\ \hline

\end{tabular}%
}
\label{tab:classifier_network}
\end{table} 

\section{Distribution of norms in the codebooks}
\label{app:distribution_of_norm}
\setcounter{figure}{0}

\begin{figure}[H]
  \centering
  \includegraphics[]{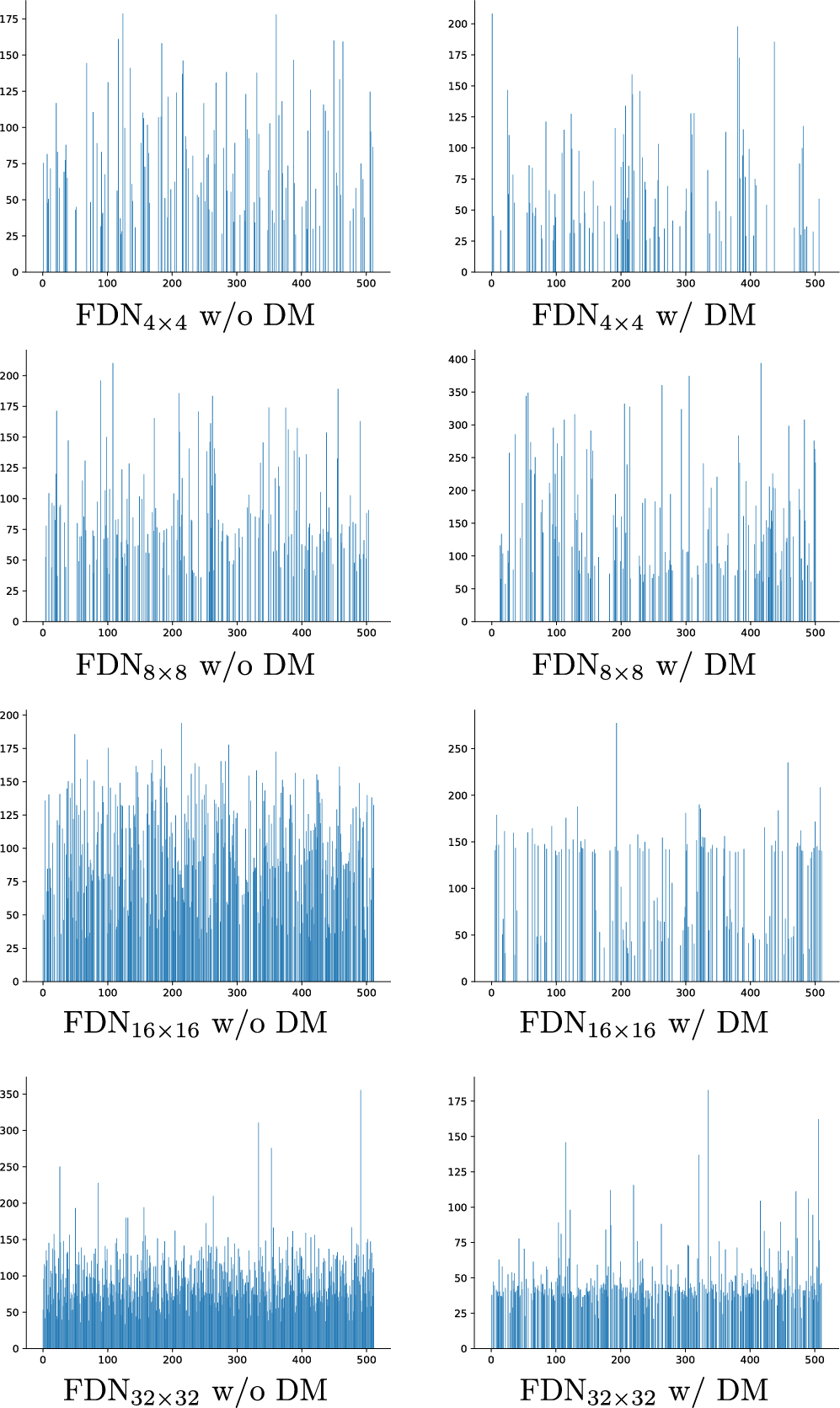}
  \caption{\textbf{Distribution of norms in the codebooks.} For each configuration with or without distribution matching (DM), the magnitude of norms (vertical axis) along with a total of 512 $(= K)$ code vectors (horizontal axis) is indicated. Note that sparsity of the code vectors was remarkable at the lower resolution of the latent space. Moreover, there is a tendency that distribution matching enhanced such sparsity in each level of the latent resolution.}
  \label{fig:codebook_norm}
\end{figure}

Distributions of norms of code vectors in the codebooks that correspond to normal anatomy codes are demonstrated in {\bf Fig. \ref{fig:codebook_norm}}. Note that, at the lower spatial resolution of the latent space, such as $4 \times 4$ and $8 \times 8$, a significant portion of the code vectors had minimum norm close to zero. Meanwhile, the number of code vectors exhibiting larger norms were increased in the resolution of $16 \times 16$ and $32 \times 32$. Furthermore, there seems to be a tendency that distribution matching induced the sparsity of the code vectors, indicating its regularization effect in the selection process of code vectors.

\section{Comparison of computational time at the distance calculation}
\label{app:comparison_of_computational_time}
\setcounter{figure}{0}   

The computational times according to Euclidean distance using the continuous codebook $\bm{e}$ and Hamming distance using the optimized binarized codebook $\bm{b}^*$ were compared. For the evaluation, for each $K (= 512)$ code vector, the distances between all $K$ code vectors, including itself, were exhaustively calculated. The Python functions for the Euclidean and Hamming distance calculations are shown in {\bf Listing \ref{distance_code}}. The calculations were repeated five times on a computer, and the average times using Euclidean distance $t_\mathrm{E}$ and Hamming distance $t_\mathrm{H}$ were calculated. The reduction ratio was reported as follows: $1 - \frac{t_\mathrm{H}}{t_\mathrm{E}}$.

\setcounter{lstlisting}{0}
\begin{lstlisting}[language=Python, caption=\textbf{Example Python code for the distance calculation.}, label=distance_code]
import numpy as np
    
def calc_hamming_distance(a, b):
    return np.count_nonzero(a!=b)
    
def calc_euclidean_distance(a, b):
    return np.linalg.norm(a - b)
\end{lstlisting}

\section{Example learning processes}
\label{app:example_learning_process}
\setcounter{figure}{0}   

\begin{figure}[H]
  \centering
  \includegraphics[]{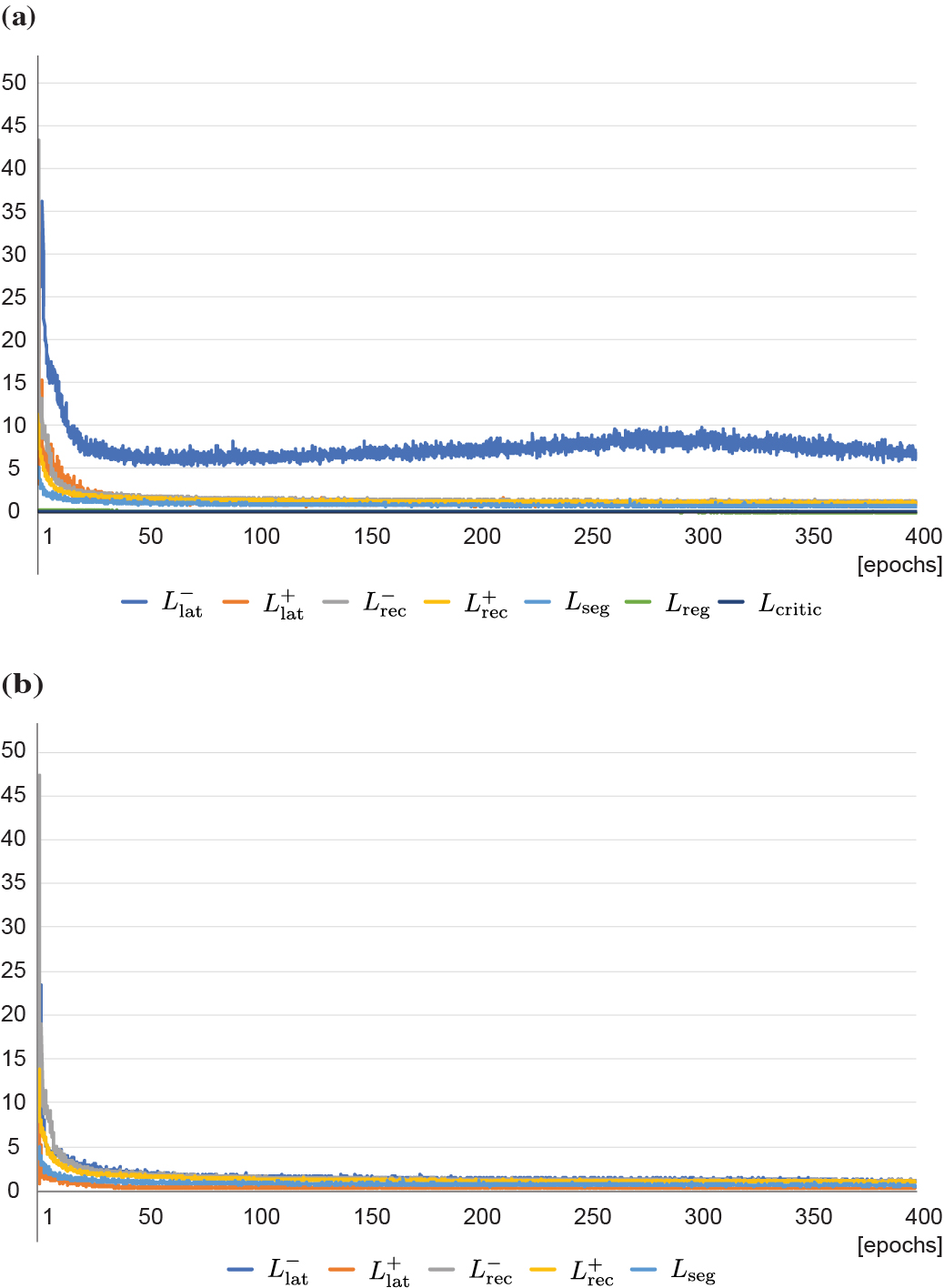}
  \caption{\textbf{Example of the learning process of the feature decomposing network.} \textbf{(a)} When using distribution matching to $\mathrm{FDN}_{8 \times 8}$, $L_\mathrm{lat}$ stopped dropping in the middle range of the value, indicating the regularization effect, owing to the distribution matching. \textbf{(b)} Without using distribution matching, $L_\mathrm{lat}$ also consistently reduced as the training proceeded.}
  \label{fig:learning_curve}
\end{figure}

The example of learning process of $\mathrm{FDN}_{8 \times 8}$ trained with distribution matching is shown in {\bf Fig. \ref{fig:learning_curve}a}. Note that almost all loss functions except for $L_\mathrm{lat}$ were reduced consistently as the training steps increased. The learning curve of $L_\mathrm{lat}$ was affected by $L_\mathrm{reg}$ that imposed some regularization effects on the selection process of normal anatomical codes, rendering it difficult for the encoder to map latent representations into positions close to code vectors. Without using distribution matching, $L_\mathrm{lat}$ could also be reduced constantly according to the training steps ({\bf Fig. \ref{fig:learning_curve}b}). 

\section*{Acknowledgement}
The authors thank the members of the Division of Medical AI Research and Development of the National Cancer Center Research Institute for their kind support. The RIKEN AIP Deep Learning Environment (RAIDEN) supercomputer system was used in this study to perform computations. 

{\bf Funding:} This work was supported by JST CREST (Grant Number JPMJCR1689), JST AIP-PRISM (Grant Number JPMJCR18Y4), and JSPS Grant-in-Aid for Scientific Research on Innovative Areas (Grant Number JP18H04908).

{\bf Competing interests:} Kazuma Kobayashi and Ryuji Hamamoto have received research funding from Fujifilm Corporation.

{\bf Author contributions:} 
\textbf{Kazuma Kobayashi: }Conceptualization, Methodology, Validation, Formal Analysis, Investigation, Resources, Data Curation, Writing---Original Draft Preparation, Writing---Review \& Editing, Visualization, Project Administration, and Funding Acquisition. \textbf{Ryuichiro Hataya: } Writing---Review \& Editing. \textbf{Yusuke Kurose: }Supervision. \textbf{Mototaka Miyake: }Data Curation. \textbf{Masamichi Takahashi: }Data Curation. \textbf{Akiko Nakagawa: }Resources and Data Curation. \textbf{Tatsuya Harada: }Supervision. \textbf{Ryuji Hamamoto: }Supervision and Funding Acquisition. All authors have read and agreed to the submitted version of the manuscript.

\bibliography{references}

\end{document}